\documentclass{article}
\PassOptionsToPackage{numbers}{natbib}
\usepackage[preprint]{neurips_2026}
\usepackage{enumitem}
\usepackage{natbib}
\usepackage[utf8]{inputenc}
\usepackage[T1]{fontenc}
\usepackage{hyperref}
\usepackage{url}
\usepackage{booktabs}
\usepackage{amsfonts}
\usepackage{nicefrac}
\usepackage{microtype}
\usepackage{xcolor}
\usepackage{amsmath, amssymb, mathtools, bm}
\usepackage{tabularx}
\usepackage[most]{tcolorbox}
\usepackage{subcaption}
\usepackage{float}
\usepackage[linesnumbered,ruled,vlined]{algorithm2e}
\SetKwInOut{KwIn}{Input}
\SetKwInOut{KwOut}{Output}
\usepackage{caption}
\usepackage{tikz}
\usetikzlibrary{positioning, arrows.meta, calc, shapes, fit}

\title{On the Geometry of Games and their Solvers}

\author{Yaqi Sun\thanks{Equal contribution} \\Queen Mary University of London\And Julian Ma$^*$\\University College London \And 
David Mguni\thanks{{Correspondence to: d.mguni@qmul.ac.uk}}
\\Queen Mary University of London
}

\date{}

\begin{document}

\maketitle

\begin{abstract}
A central challenge in game theory and learning systems such as GANs is understanding which algorithms can efficiently compute equilibria across the heterogeneous landscape of games. Equilibrium computation is typically studied solver by solver and game class by game class, yielding strong local guarantees but a fragmented view of solver behaviour. Existing discrete taxonomies often provide an incomplete account of where algorithms succeed. We study this problem through a solver-game map linking games to effective solver dynamics. Classical theory identifies isolated regions of this map but provides limited insight into intermediate or overlapping regimes, suggesting that solvability is governed by latent structural properties defining a continuous solver-aligned geometry of games. We formalise this perspective through structure-aware solver synthesis. A learned structure recogniser maps each game to a low-dimensional solver-aligned representation, and a policy maps this representation to effective primitive mechanisms, adapting solver behaviour across regimes. This reveals regions where particular solver dynamics are effective and where mixtures of primitives are required rather than a single dominant solver. A bounded residual acts as a local corrector and diagnostic signal for incomplete solver bases or representations. The framework yields both an adaptive solver and an analytical lens: games with similar optimisation dynamics cluster together, revealing continuous regions of algorithmic validity and overlapping solver behaviour. Empirically, we show that fixed primitives exhibit systematic regime mismatch, while the learned representation organises game space into a structured cartography aligned with solver behaviour. These results suggest viewing equilibrium computation as the joint problem of learning solver mechanisms and mapping the geometry of solvability.
\end{abstract}

\section{Introduction}

Equilibrium computation is a basic problem in game theory and modern learning systems such as GANs, where training dynamics can be viewed as strategic interactions between competing objectives~\cite{deng2023complexity,chen2006settling, li2024survey}. Existing algorithmic theory is largely organised solver by solver and game class by game class~\cite{yang2020game, papadimitriou2005computing}. Gradient-based dynamics are analysed in potential or descent-structured regimes~\cite{balduzzi2018mechanics}; extra-gradient and optimistic methods are developed for rotational or zero-sum settings~\cite{antonakopoulos2020adaptive, daskalakis2018limit, mokhtari2020unified}; and response-based dynamics are studied in selected discrete classes~\cite{leslie2020best, swenson2018best}. This gives strong local guarantees, but leaves a fragmented account of solver behaviour beyond these canonical regimes.

In practice, game instances rarely fall cleanly into a single class. They may interpolate between potential, harmonic, zero-sum, symmetric, and degenerate structure~\cite{chen2022convergence, candogan2013near}. In such intermediate regimes, solver performance can change sharply: a method that is effective in one region may become unstable or inefficient nearby, and the best-performing mechanism may vary with the underlying structure~\cite{chen2022convergence, sparrow2008fictitious}. As a result, discrete taxonomies provide only a partial guide to algorithm selection. This limits not only which solver we choose, but also how we interpret why a solver succeeds or fails on a given game.

This raises a fundamental question: \emph{how do algorithmic mechanisms relate to game structure, and which mechanisms enable efficient solution across heterogeneous game landscapes?}

We study this question through a \emph{solver--game map} that links game instances to the solver dynamics that are effective on them. Classical theory can be viewed as identifying isolated regions of this map, corresponding to convergence guarantees for particular algorithms on particular classes. Our aim is to model the regions between these cases. We therefore consider a continuous, solver-aligned geometry of games, where proximity is determined not only by game descriptors or family labels, but by the optimisation dynamics that games admit.

\begin{figure}[t]
    \centering
    \includegraphics[width=.8\linewidth, height=4.1cm]{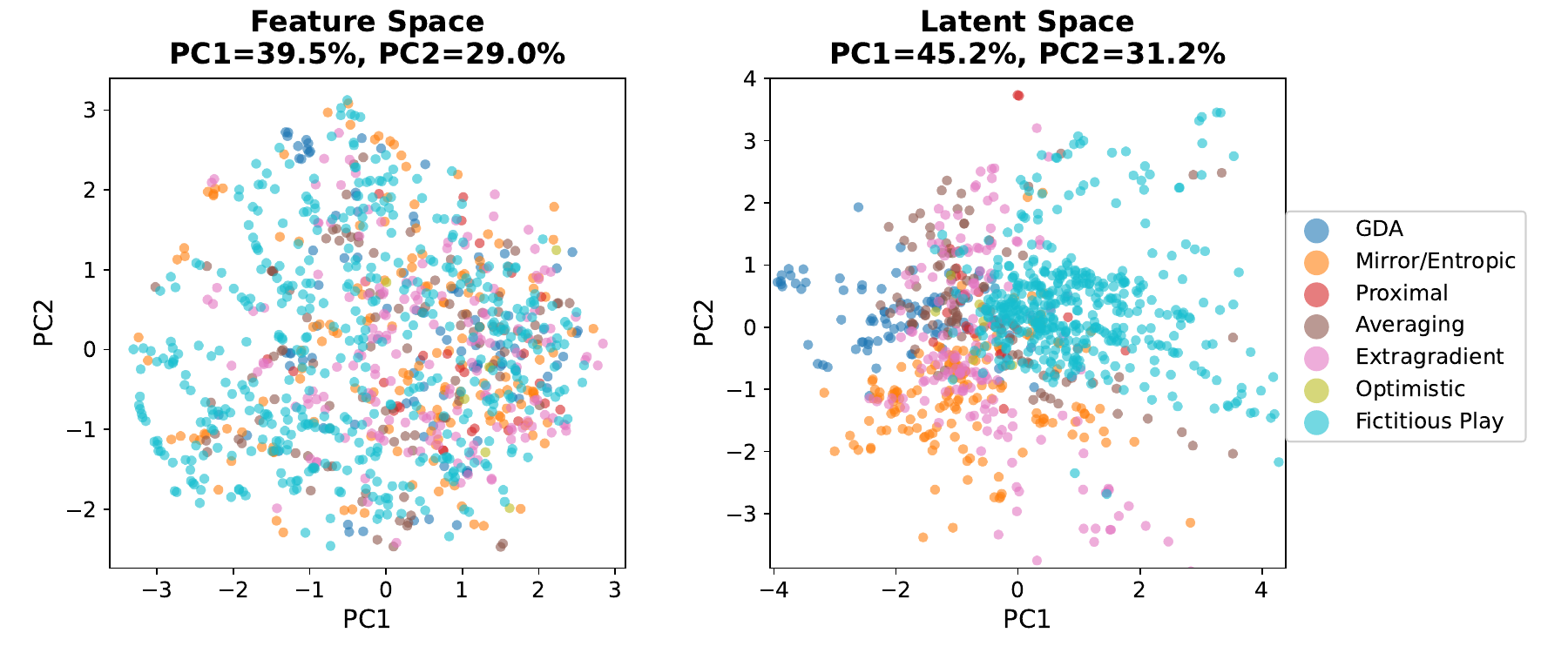}
    \caption{
    From fragmented solver behaviour to solver-aligned geometry.
    Each point denotes one game instance, and its colour indicates the primitive solver with the lowest exploitability AUC on that game.
    The horizontal and vertical axes are the first two principal components obtained by PCA.
    \textbf{Left:} PCA projection of the raw feature space, defined by the hand-designed game diagnostics before they are passed to the structure recogniser; solver labels are highly intermingled.
    \textbf{Right:} PCA projection of the learned latent space, constructed from the recogniser-generated representation \(\hat z\); games with similar solver efficacy form more coherent regions.
    }
    \label{fig:intro-cartography} \vspace{-.5cm}
\end{figure}

Figure~\ref{fig:intro-cartography} illustrates this perspective. In both panels, the horizontal and vertical axes are the first two principal components, PC1 and PC2, obtained by applying PCA to the corresponding representation space; the percentages in parentheses indicate the fraction of total variance explained by each component.
In the raw diagnostic feature space, solver success appears fragmented: games for which different primitives achieve the lowest exploitability AUC are highly intermingled, suggesting that the hand-designed diagnostics alone do not provide a solver-aligned coordinate system.
In contrast, after passing the game features through our structure recogniser, the learned representation \(\hat z\) induces a latent space in which games with similar solver efficacy form more coherent regions.
This reveals a coherent cartography of game space in which solver behaviour defines the underlying geometry.

Formally, given a game $G=(A,B)$, we introduce a structure recogniser that produces a low-dimensional representation $\hat z(G)$ learned from solver-relevant signals. A solver policy maps this representation to a convex mixture of primitive update mechanisms, including gradient, mirror-style~\cite{beck2003mirror}, extra-gradient~\cite{antonakopoulos2020adaptive}, optimistic~\cite{daskalakis2018limit}, and response-based dynamics~\cite{matsui1992best}. The resulting prior update lies in the convex hull of known solver behaviours, allowing the method to interpolate among mechanisms rather than select a single primitive. A bounded residual provides a local correction beyond this hull and serves as a diagnostic signal for regions where the primitive basis or representation is insufficient.

Empirically, we find that fixed primitives exhibit strong regime-dependent mismatch. The learned representation organises games into coherent solver-aligned regions: high-margin regions admit stable primitive identities, while transition regions favour mixtures of primitives. Residual activation highlights cases where the primitive hull is locally insufficient. These results support the view that equilibrium computation can be studied as the joint problem of learning solver mechanisms and mapping the geometry of solvability.

\paragraph{Contributions}
\begin{itemize}[leftmargin=*, noitemsep]

\item We identify a structural limitation in the prevailing solver-by-solver, class-by-class paradigm, which yields a fragmented understanding of solver behaviour across heterogeneous game space.

\item We introduce the notion of a \emph{solver--game map} and propose a solver-induced geometric view in which games are organised by the optimisation dynamics they admit rather than by discrete game classes alone.

\item We develop a \emph{structure-aware solver synthesis} framework that maps inferred game structure to convex mixtures of primitive solver mechanisms through a learned solver-aligned representation.

\item We show empirically that no single primitive solver dominates across the game landscape, and that soft mixtures consistently outperform hard solver selection in low-margin transition regions where solver identities are unstable.

\item We introduce a bounded residual as a local corrector and diagnostic signal for regions where the primitive basis or structural representation is insufficient.

\item We show empirically that solver-aligned representations outperform decomposition-based proxies and raw generative coefficients, suggesting that the geometry governing solvability is richer than existing analytical game coordinates.

\end{itemize}

\textbf{Worked example.}
%
Consider the interpolated \(3\times 3\) game
\[
A_\lambda = (1-\lambda)A_{\mathrm{RPS}} + \lambda A_{\mathrm{pot}}, 
\qquad
B_\lambda = -(1-\lambda)A_{\mathrm{RPS}} + \lambda A_{\mathrm{pot}},
\quad \lambda \in [0,1],
\]
where \(A_{\mathrm{RPS}}\) is the Rock--Paper--Scissors payoff and \(A_{\mathrm{pot}}\) is a simple potential game. At \(\lambda=0\), the game is zero-sum and exhibits rotational dynamics; at \(\lambda=1\), it is potential-like and admits descent structure. For intermediate \(\lambda\), the game lies between canonical classes. In such regimes, there is no reliable predictive rule for selecting an appropriate solver: small changes in \(\lambda\) can alter the effectiveness of gradient, extra-gradient, or response-based dynamics, and classical classifications provide little predictive guidance.
In our framework, the recogniser maps each instance to a solver-relevant representation
\[
\hat z(G) \approx 
(\hat z_{\mathrm{pot}}, \hat z_{\mathrm{harm}}, \hat z_{\mathrm{zs}}, \hat z_{\mathrm{sym}}, a_{\mathrm{mono}}),
\]
which varies smoothly with \(\lambda\). For intermediate values, \(\hat z\) typically reflects mixed structure (e.g.\ non-negligible potential and rotational components), rather than a single dominant class.
The solver is then constructed as a mixture of primitive mechanisms conditioned on \(\hat z\), placing mass on both descent-aligned and anti-cycling primitives in transition regimes. The residual remains small in well-explained regions and activates only when the primitive mixture is locally insufficient, acting as a corrective signal and a diagnostic of missing structure.
This example illustrates the central point of the paper: interpolated games do not admit a stable solver identity under class-based reasoning. Instead, they are naturally described by a continuous, solver-aligned representation, under which solver behaviour varies smoothly and mixtures replace hard algorithm selection.

Table~\ref{tab:primitive-convergence-map} summarises the key structural pattern in the literature: convergence guarantees are not attached to a single universal solver, but to specific solver mechanisms acting within specific game classes. Fictitious play is known to converge in zero-sum games, \(2\times 2\) games, and potential games, but not in general non-zero-sum games~\cite{krishna1998convergence}; best-response and better-response dynamics are effective in potential and acyclic settings; no-regret dynamics are much more robust across general games, but typically only guarantee convergence to correlated or coarse correlated equilibrium rather than Nash equilibrium. This patchwork of results motivates our primitive-based design: instead of selecting a single solver a priori, we treat these mechanisms as building blocks and learn how to combine them as a function of game structure. 

\section{Related Work}
\label{sec:related}

Our work connects three strands of literature. First, game decomposition provides analytical coordinates for understanding strategic structure~\cite{candogan2011flows, chen2022convergence}. Potential games formalise classes in which incentives align with a global potential \citep{monderer1996potential, la2016potential}, while the potential--harmonic--nonstrategic decomposition gives a canonical structural analysis of finite games \citep{candogan2011flows}. These tools motivate our diagnostics, but our empirical results suggest that decomposition coordinates alone are not sufficient to predict solver behaviour.
Second, equilibrium computation in games has largely developed solver-by-solver and class-by-class~\cite{monderer1996fictitious, swenson2018best}. Mirror-Prox and related extra-gradient methods provide guarantees for monotone variational inequalities and convex--concave saddle-point problems \citep{nemirovski2004prox}. In adversarial learning, optimistic and extrapolation-based methods address cycling and instability in GAN-style games \citep{daskalakis2018training,gidel2019variational}. Differentiable game dynamics have also been studied through structural decompositions of the game Jacobian, separating potential-like and Hamiltonian components \citep{balduzzi2018mechanics}. Our work differs by learning a solver-aligned geometry from observed primitive performance rather than assuming that a fixed analytical decomposition is the correct coordinate system.
Third, learned optimisation casts the design of update rules as a learning problem \citep{andrychowicz2016learning}. Our setting is different: the goal is not a generic optimiser, but a game solver whose behaviour is conditioned on a learned representation of game structure. This lets the learned policy serve both as an adaptive solver and as a tool for mapping the geometry of solvability.

\section{Problem Setting and Notation}
\label{sec:problem}

We consider two-player games with payoff matrices
\(
A,B \in \mathbb{R}^{n \times m},
\)
and mixed strategies
\(
x \in \Delta^n, y \in \Delta^m
\),
where \(\Delta^d\) denotes the probability simplex. The objective is to compute or approximate a Nash equilibrium, or equivalently to reduce exploitability or duality gap below a prescribed tolerance. The solver operates iteratively. At iteration \(t\), it maintains the current state
\(
\theta_t := (x_t,y_t),
\)
and applies an update rule
\(
\theta_{t+1} = \mathcal U_\Phi(\theta_t; A,B,u_t),
\)
where \(u_t\) denotes optimisation diagnostics such as step index, exploitability proxies, gradient norms, primitive alignments, or cycling indicators. For a solver \(F\), we write \(\mathrm{Err}_T(F;G)\) for the terminal error after \(T\) iterations on game \(G\), using exploitability as the default error measure. For a tolerance \(\varepsilon>0\), we define the corresponding \(\varepsilon\)-validity region by
\[
\Gamma_\varepsilon(F) = \{G : \mathrm{Err}_T(F;G)\le \varepsilon\}.
\]

Let \(g_i\) denote a primitive solver, and let \((x_T^{(i)},y_T^{(i)})\) be the strategies obtained after \(T\) iterations on game \(G=(A,B)\). We define the performance of \(g_i\) via exploitability,
\begin{equation}
\mathrm{Err}(g_i;G)
=
\max_{x' \in \Delta^n} x'^\top A y_T^{(i)}
-
(x_T^{(i)})^\top A y_T^{(i)}
+
\max_{y' \in \Delta^m} (x_T^{(i)})^\top B y'
-
(x_T^{(i)})^\top B y_T^{(i)}.
\label{eq:primitive-error}
\end{equation}
We write \(\ell_i(G)=\mathrm{Err}(g_i;G)\) for brevity.

\section{Framework Overview}
\label{sec:framework}

\begin{figure}[t]
    \centering
    \includegraphics[width=.8\linewidth, height=4.7cm]{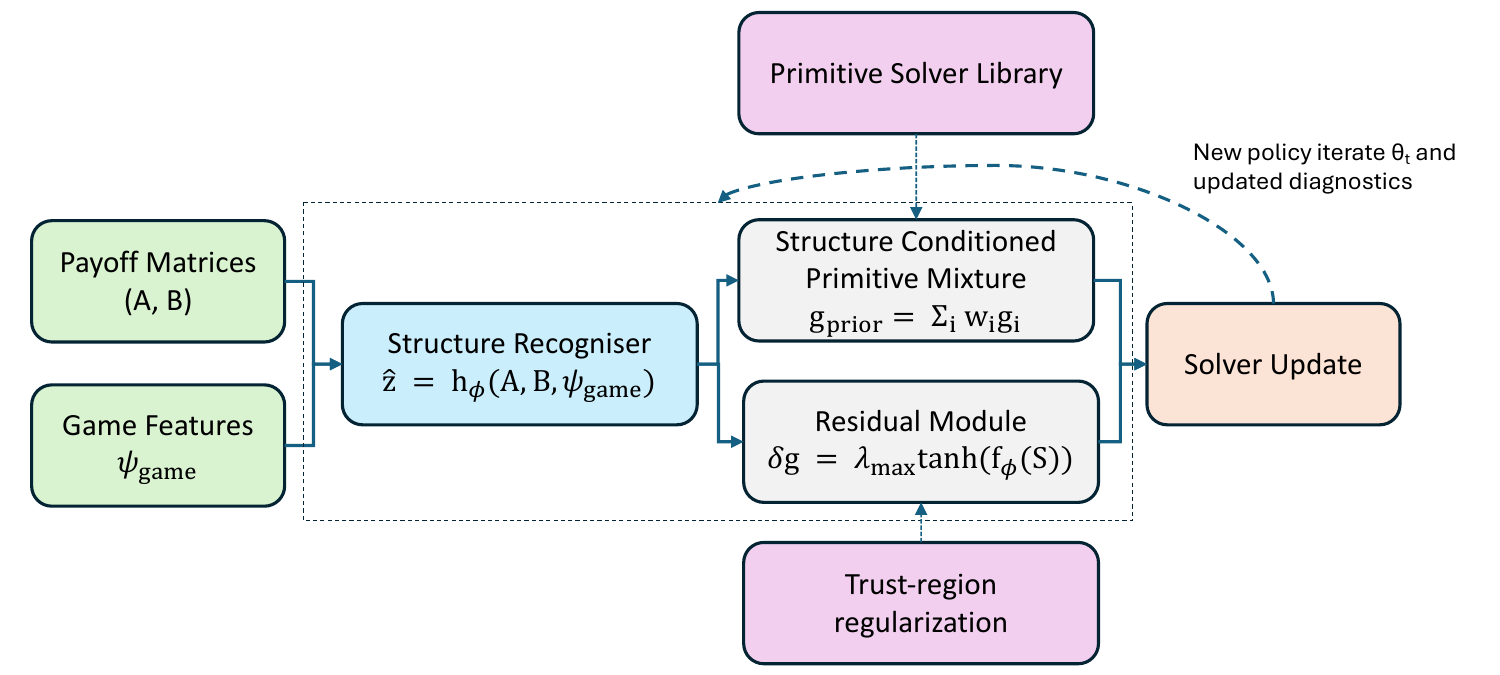}
    \caption{Overview of the proposed structure-aware solver synthesis framework.}
    \label{fig:overview} \vspace{-.5cm}
\end{figure}

The proposed method is a structure-aware end-to-end solver architecture with four coupled components.
The first component is a \emph{structure recogniser}. Given a game instance, it produces a low-dimensional solver-aligned representation \(\hat z\). This representation is not trained to reproduce a classical decomposition; instead, it is learned through downstream solver performance and later interpreted using realised game-theoretic diagnostics. This representation is not intended to be an exact label, but rather a continuous coordinate system over game space.
\newline The second component is a \emph{primitive solver module}. It evaluates a small library of update directions corresponding to canonical game-solving mechanisms, including gradient, entropic or mirror-style, extra-gradient, optimistic, best-response, and fictitious-play primitives.
\newline 
The third component is a \emph{solver synthesiser}. It forms a convex combination of primitive directions using weights conditioned on the learned representation \(\hat z\). This defines a prior solver \(g_t^{\mathrm{prior}}\) lying in the convex hull of known solver behaviours.
\newline 
The fourth component is a \emph{residual correction module}. It produces a bounded perturbation \(\delta g_t\) that allows the synthesised solver to move beyond the primitive hull when needed. A relative trust-region penalty~\cite{schulman2015trust} ensures that the residual remains a controlled refinement rather than replacing the primitive basis altogether. A small trust-style regularisation stabilises the residual but does not otherwise determine solver behaviour.
Overall, the architecture implements the map
\[
(A,B,\theta_t,u_t)
\;\longrightarrow\;
\hat z
\;\longrightarrow\;
\{g_i(\theta_t)\}_{i=1}^M
\;\longrightarrow\;
g_t^{\mathrm{prior}}
\;\longrightarrow\;
g_t^{\mathrm{synth}}
\;\longrightarrow\;
\theta_{t+1}.
\]

The framework serves both as an adaptive solver construction method and as a mechanism for inducing a solver-aligned geometry on game space. Beyond adaptive solver construction, the learned representation organises games according to observed solver behaviour through the induced primitive mixtures and residual corrections.

\section{Solver-Aligned Structure Representation}
\label{sec:structure}

Finite games admit useful decompositions into potential, harmonic, nonstrategic, zero-sum-equivalent, and symmetry-related components. These decompositions provide valuable analytical diagnostics, but they are not designed to be sufficient coordinates for predicting solver behaviour across heterogeneous regimes. A central empirical finding of this paper is precisely that replacing the learned representation with decomposition-based proxies, or with the raw generative mixture coefficients used to construct games, significantly degrades solver performance. This suggests that the geometry of solvability is related to, but not exhausted by, classical game-theoretic structure. We therefore learn a low-dimensional solver-aligned representation
\begin{equation}
\hat z = h_\Phi(A,B,\psi_{\mathrm{game}})\in\mathbb R^d,
\label{eq:structure-representation}
\end{equation}
where \(d\) is small and \(\hat z\) is trained through its downstream effect on primitive weights and rollout performance, rather than through a standalone decomposition-classification objective. The representation is trained through downstream solver performance rather than decomposition recovery. Its purpose is to organise games according to solver behaviour and support structure-conditioned routing.

Additional optimisation diagnostics used for residual geometry and analysis are described in the Appendix.

\section{Primitive Solver Basis and Residual Discovery}
\label{sec:solver}


We define a library of primitive update directions \(\{g_i\}_{i=1}^M\), each corresponding to a canonical solver mechanism. Potential-aligned primitives include gradient and entropic or mirror-style updates. Harmonic or cyclic primitives include extra-gradient and optimistic updates. Response-based primitives include best-response and fictitious-play updates. The primitive basis is therefore aligned with decomposition components of the game, rather than being an arbitrary collection of optimisation rules. For instance, the basic gradient primitive is
\[
g_x^{\mathrm{grad}} = Ay, \qquad g_y^{\mathrm{grad}} = B^\top x,
\]
the entropic primitive takes the form
\[
g_x^{\mathrm{ent}} = Ay - \tau_x \nabla H(x), \qquad
g_y^{\mathrm{ent}} = B^\top x - \tau_y \nabla H(y),
\]
the extra-gradient primitive is defined through a lookahead step, the optimistic primitive uses a momentum-like correction across consecutive gradients, and the best-response and fictitious-play primitives use either instantaneous or empirical response structure. We defer the full list of equations to the appendix.
%
%
Given the primitive basis, the prior solver is defined as a convex combination
\begin{align}
g^{\mathrm{prior}}_t
=
\sum_{i=1}^M w_i(\hat z;\Phi)\, g_i(\theta_t),
\qquad
w_i(\hat z;\Phi)\ge 0,
\qquad
\sum_{i=1}^M w_i(\hat z;\Phi)=1.
\label{eq:primitive-mixture}
\end{align}
which may be equivalently written as \(
g^{\mathrm{prior}}_{x,t}
=
\sum_{i=1}^M w_i(\hat z;\Phi)\, g_{x,i}(\theta_t),
\qquad
g^{\mathrm{prior}}_{y,t}
=
\sum_{i=1}^M w_i(\hat z;\Phi)\, g_{y,i}(\theta_t)
\). In the current implementation, \(w(\hat z;\Phi)\) is static within a solver rollout: one primitive mixture is selected for a game and then used across the iterative trajectory. The policy outputs primitive logits \(r_\Phi(\hat z)\), from which we define both a soft mixture
\[
w_{\mathrm{soft}}(G)
=
\mathrm{softmax}\!\big(\beta r_\Phi(\hat z(G))\big)
\]
and a hard top-1 primitive
\[
w_{\mathrm{hard}}(G)
=
e_{\arg\max_i r_{\Phi,i}(\hat z(G))}.
\]
The soft mixture is used for mixture-based solver synthesis, while the hard vector is used to evaluate whether the learned policy recovers the best pure primitive. This distinction is important because soft mixtures may perform well even when the top-1 primitive is incorrect, especially in regions where several primitives have similar exploitability.
The dependence of \(w\) on \(\hat z\) is the primary mechanism by which game structure influences solver choice. However, we do not assume that each region of game space has a single sharply dominant primitive. Empirically, many regions are ambiguous: the best and second-best primitives may have very similar AUC or terminal exploitability. We therefore interpret the learned mixture as a robust solver policy over primitive mechanisms, and interpret hard primitive territories only when the performance margin is sufficiently large.

\subsection{Residual solver discovery}

To move beyond the convex hull of the primitive basis, we introduce a bounded residual update generated from solver-geometry features. Let
\(
\mathcal S_t
=
\mathcal S(G,u_t,\hat z,w,g_t^{\mathrm{prior}})\)
denote features containing the recognised structure, diagnostics, primitive weights, primitive proposal norms, prior-step norms, and relative geometry terms. The residual module outputs
\begin{equation}
\delta g_t
=
\lambda_{\max}\tanh\!\big(f_\Phi(\mathcal S_t)\big).
\label{eq:residual-update}
\end{equation}
A scalar gate controls whether the residual is active:
\begin{equation}
q_\Phi(G)
=
\sigma(h_\Phi(\hat z(G),s(G)))\in[0,1],
\end{equation}
where \(s(G)\) denotes optional static game statistics or hardness features. The synthesised update is
\begin{equation}
g^{\mathrm{synth}}_t
=
g^{\mathrm{prior}}_t
+
q_\Phi(G)\delta g_t.
\label{eq:synthesised-update}
\end{equation}

This design makes the residual a local corrector rather than a replacement for the primitive mixture. In our experiments, the residual is most useful in local regions where the primitive hull is close to adequate but leaves systematic error. Persistent residual activation therefore acts as a diagnostic: it indicates either a missing primitive, a missing structural coordinate, or an interaction between known coordinates that is not captured by the current representation. A small trust-region style regularisation is applied to the residual to ensure stable updates, but does not otherwise determine solver behaviour.

\section{Training Method}

\textbf{Structure recogniser and weight prediction network.}
The structure recogniser is the first stage of the architecture. It takes as input the game payoff matrix together with auxiliary game-level features, and maps them to a five-dimensional continuous structure vector
\(
\hat z = h_\Phi(A,B,\psi_{\mathrm{game}})\in[0,1]^5.
\)
This representation is then passed to a weight prediction network, which outputs the mixture weights
\(
w(G)=(w_1(G),\dots,w_M(G)),
\)
with one weight assigned to each pure primitive solver in the library. The coordinates of \(\hat z\) are not constrained to form a simplex and are not treated as mutually exclusive structural labels. Rather, they provide a low-dimensional control signal for generating the primitive weights \(w(G)\). The recogniser is deliberately not trained to reproduce decomposition labels. Instead, \(\hat z\) is learned through its downstream effect on predicted primitive weights and solver performance, measured by AUC over the rollout.
\newline
\textbf{Phase I: one-hot supervision for weight initialisation.}
\textcolor{black}{We train the recogniser and weight prediction network jointly by minimising the KL divergence between the predicted weight distribution and this oracle target:}
\textcolor{black}{\(
\mathcal L_{\mathrm{Phase\,I}}
=
\mathbb E_G
\big[
D_{\mathrm{KL}}(w^\star(G)\,\|\,w(G))
\big].
\)
}
We then train the weight prediction network against this target. Under this one-hot restriction, the target weights are available as ground truth, since they are obtained directly by evaluating the primitive library on each game and selecting the primitive with lowest AUC.
\newline
\textbf{Phase II: continuous end-to-end training.}
For a predicted weight vector \(w(G)\), we execute the induced mixed solver for \(T\) steps through a differentiable computation graph and measure performance using the area under the exploitability trajectory. We optimise a normalised AUC objective
\textcolor{black}{
\[
\mathcal L_{\mathrm{rollout}}
=
\frac{
\mathrm{AUC}(w;G)
}{
\mathrm{AUC}_{\mathrm{best}}(G)+\varepsilon
},
\]
where $\mathrm{AUC}_{\mathrm{best}}(G) = \min_i \mathrm{AUC}(g_i;G)$ is the best pure-primitive performance on game $G$. This oracle-relative normalisation ensures that gradient magnitudes are comparable across easy and hard games without introducing the instability of min-max scaling.
To prevent routing collapse during end-to-end training, we retain a behavioural anchor that keeps the learned routing close to a soft target derived from primitive-wise performance:
\[
b(G)
=
\mathrm{softmax}\!\left(-\ell(G)/\tau_{\mathrm{beh}}\right),
\qquad
\mathcal L_{\mathrm{behav}}
=
D_{\mathrm{KL}}(b(G)\,\|\,w(G)).
\]
An entropy regulariser discourages premature specialisation:
\(
\mathcal L_{\mathrm{ent}} = -H(w(G)) = \sum_{i=1}^M w_i(G)\log w_i(G).
\)
The full Phase~II objective is:
\[
\mathcal L_{\mathrm{Phase\,II}}
=
\mathcal L_{\mathrm{rollout}}
+
\lambda_{\mathrm{behav}}\,\mathcal L_{\mathrm{behav}}
+
\lambda_{\mathrm{ent}}\,\mathcal L_{\mathrm{ent}}.
\]
Both the recogniser and weight prediction network receive gradients through all terms, allowing the structure representation to adapt to the end-to-end solver-performance signal.
}
\newline
\textbf{Training schedule.}
Training proceeds in two phases: oracle-initialised routing followed by differentiable rollout optimisation with continuous primitive mixtures. Phase II uses loss-based prioritised sampling with an exponential moving average of rollout loss and a short uniform-sampling warmup.

\textbf{Residual as a probe of missing structure.} Persistent residual activation indicates regions where the primitive basis or representation is insufficient (see Appendix \ref{app:residual_probe} for more details).

\section{Experiments}
\label{sec:experiments}

\begin{figure}[t]
    \centering

    \setlength{\abovecaptionskip}{2pt}
    \setlength{\belowcaptionskip}{-3pt}

    \begin{subfigure}[t]{0.48\linewidth}
        \centering
        \includegraphics[
            height=4.2cm,
            trim={4 6 12 6},clip
        ]{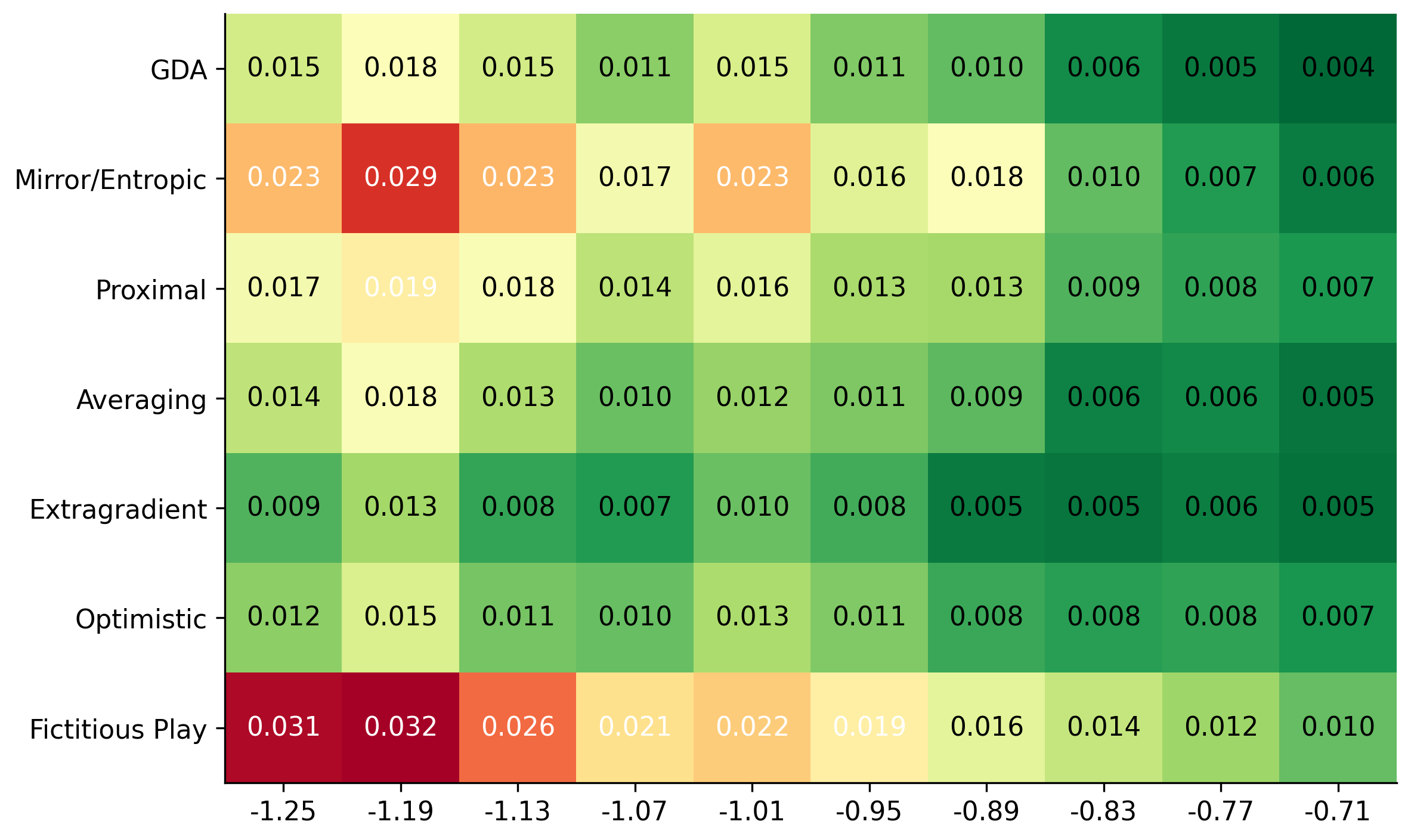}
        \caption{\footnotesize Mean exploitability AUC.}
        \label{fig:a-mono-auc}
    \end{subfigure}
    \hfill
    \begin{subfigure}[t]{0.48\linewidth}
        \centering
        \includegraphics[
            height=4.2cm,
            trim={4 6 12 6},clip
        ]{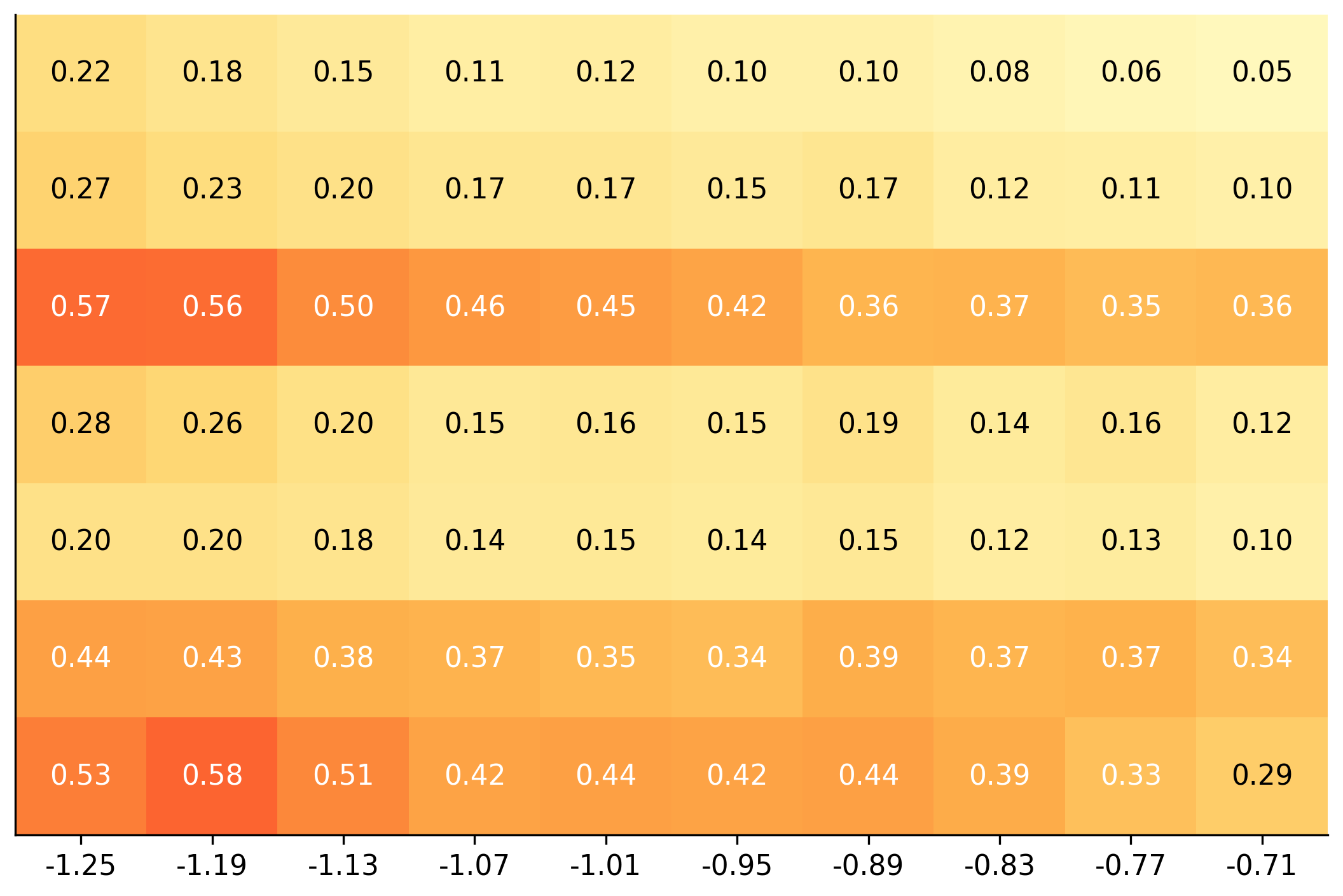}
        \caption{\footnotesize Failure fraction.}
        \label{fig:a-mono-failure}
    \end{subfigure}

    \vspace{5pt}

    \caption{
    Solver behaviour across the game-structure spectrum (\(N=35{,}804\)).
    \textbf{Left}: average performance varies non-uniformly with \(a_{\mathrm{mono}}\).
    \textbf{Right}: failure rates vary substantially by primitive and structure bin, showing that solver choice is structure-dependent.
    }
    \label{fig:a-mono-spectrum}

    \vspace{-6pt}
\end{figure}

\begin{figure}[t]
    \centering

    \begin{subfigure}[t]{0.34\linewidth}
        \centering
        \includegraphics[width=\linewidth]{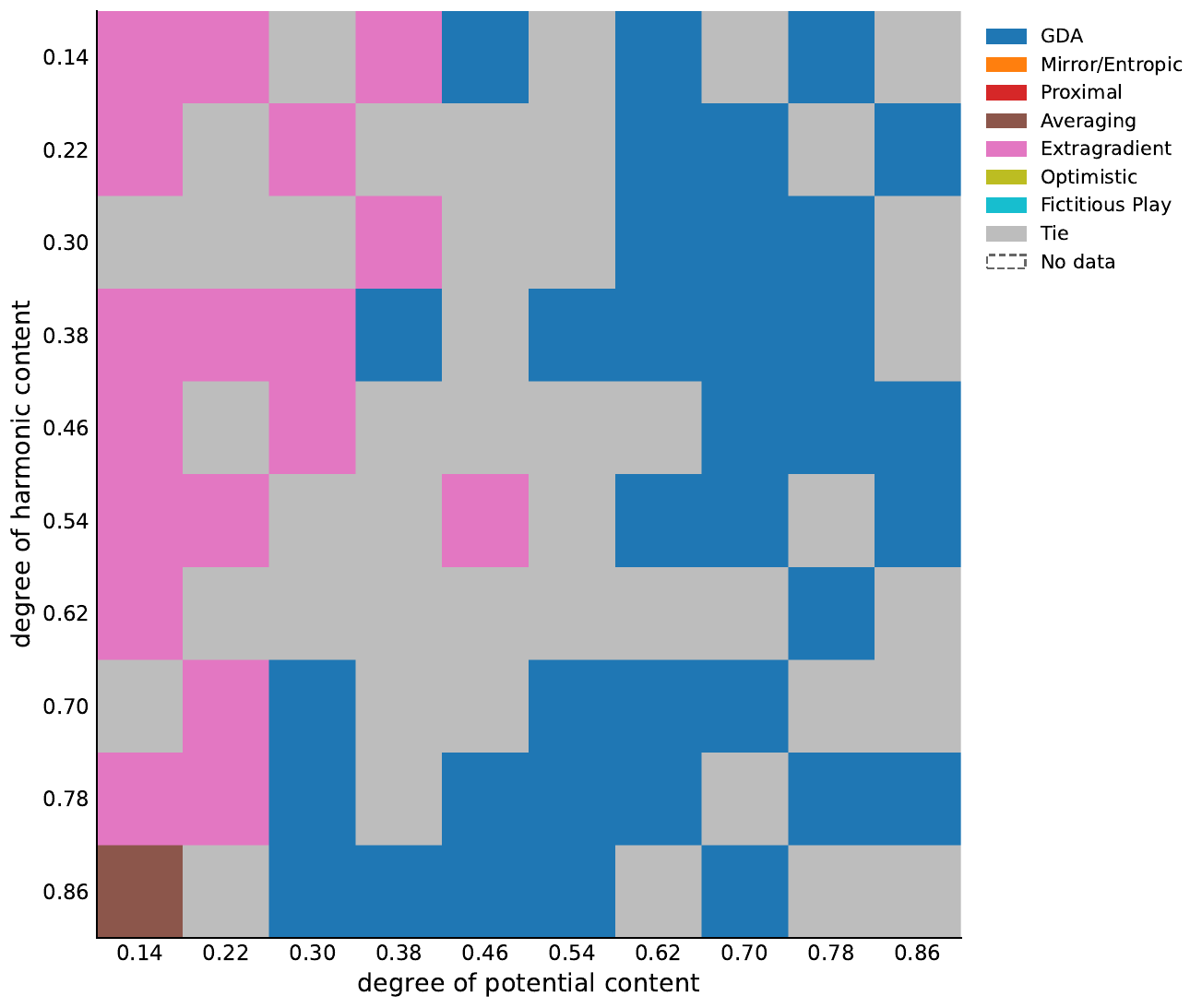}
        \caption{\footnotesize $z_{\mathrm{harm}} \times z_{\mathrm{pot}}$.}
        \label{fig:winner-heatmap-zpot-zharm}
    \end{subfigure}
    \hfill
    \begin{subfigure}[t]{0.34\linewidth}
        \centering
        \includegraphics[width=\linewidth]{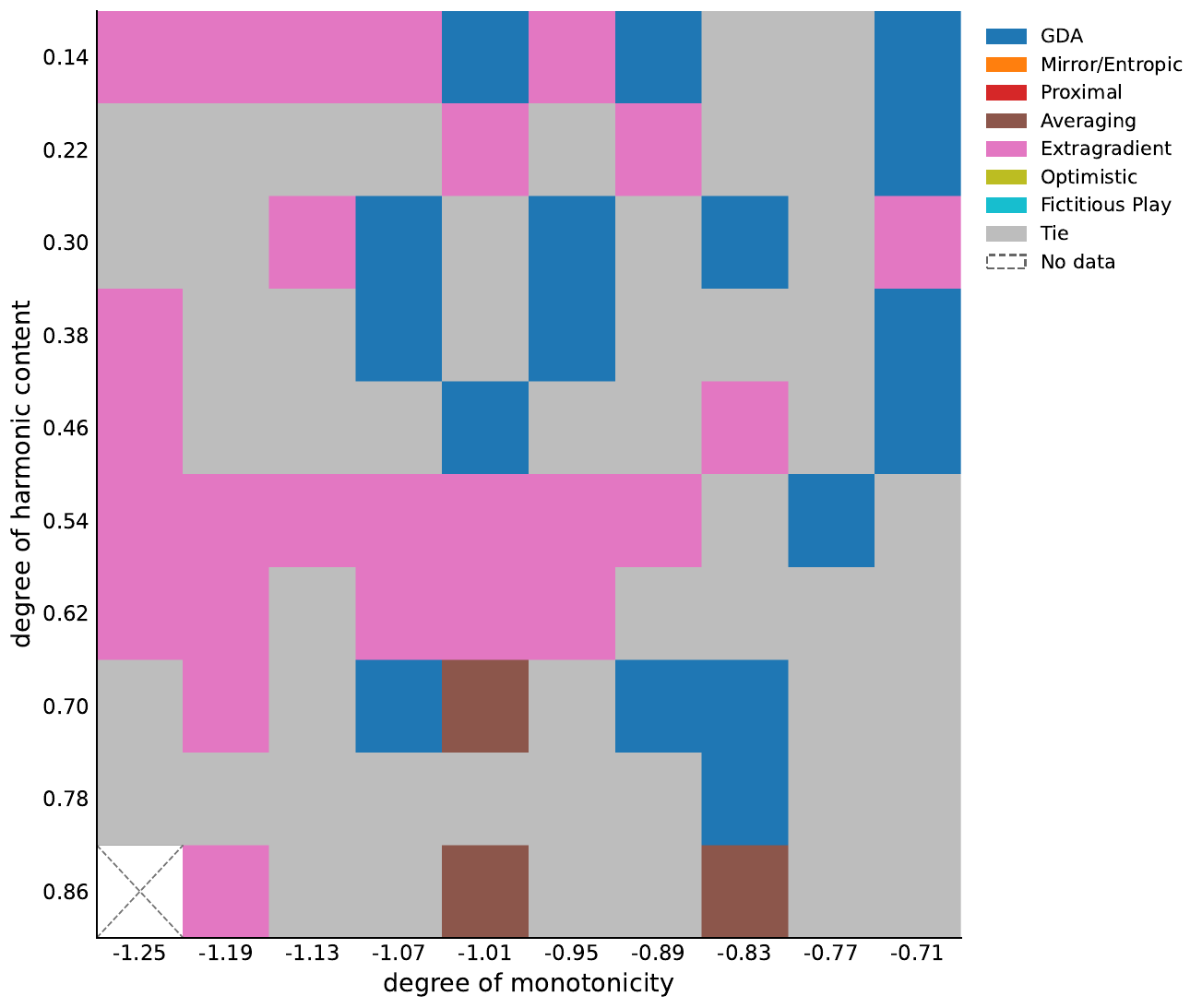}
        \caption{\footnotesize $a_{\mathrm{mono}} \times z_{\mathrm{harm}}$.}
        \label{fig:winner-heatmap-amono-zharm}
    \end{subfigure}
\vspace{1em}
    \caption{
    Oracle-best primitive by diagnostic bin.
    Each heat map shows regions where different primitives dominate.
    Greyed out regions denote bins where the performance difference between the best two primitives is not statistically significant.
    }
    \label{fig:winner-heatmaps}

    \vspace{-5mm}
\end{figure}

\textbf{Experimental overview and motivation.}
Our empirical evaluation asks whether the proposed framework can translate heterogeneous game structure into solver-adaptive optimisation. We first quantify solver mismatch by showing that no fixed primitive is robust across the full game corpus (\S\ref{subsec:exp-mismatch}), then test whether this mismatch admits a coherent solver-aligned latent geometry (\S\ref{subsec:exp-latent}). We finally evaluate the learned compositional solver against fixed, uniform, routed, and oracle baselines (\S\ref{subsec:exp-synthesis}), with ablations and residual diagnostics isolating the sources and limits of the learned representation (\S\ref{sec:ablations}; \S\ref{subsec:exp-residual}).

\textbf{Dataset.}
The generation process is described in Appendix~\ref{sec:data_construction}. 
We study two-player matrix games spanning zero-sum, potential-like, harmonic, symmetric, interpolated, and perturbed regimes. 
These regimes are chosen to cover qualitatively different optimisation geometries, including descent-like, rotational, mixed, and boundary cases.

\textbf{Experimental setting.}
Performance is evaluated using exploitability AUC over solver trajectories, averaged across three random seeds. 
Lower AUC indicates that a solver reduces exploitability faster and more reliably over the rollout horizon.

\textbf{Baselines.}
We compare against standard primitive solvers: GDA~\cite{daskalakis2018limit}, Mirror Descent~\cite{beck2003mirror}, Extra-gradient~\cite{antonakopoulos2020adaptive}, Optimistic methods~\cite{daskalakis2018limit}, Fictitious Play~\cite{berger2007brown}, Best-response~\cite{swenson2018best}, and Averaging. 
These baselines are chosen because they represent complementary update mechanisms for different game structures: simultaneous gradient learning, geometry-aware mirror updates, lookahead or optimistic corrections for rotational dynamics, response-based adaptation, and simple averaging-based stabilisation. 
Since our method synthesises updates from the same primitive library, these comparisons directly test whether the learned framework does more than choose a single known solver.

We also include three reference policies. 
The best fixed primitive tests whether one globally chosen solver is sufficient across the game distribution. 
The equal-weight mixture tests whether naive averaging of primitives is enough without structure-conditioned routing. 
The per-game oracle primitive selects the best primitive in hindsight for each game and serves as a non-deployable upper bound for hard primitive selection.

\subsection{The cost of solver mismatch}
\label{subsec:exp-mismatch}

We first evaluate how fixed primitives behave across a large corpus of heterogeneous games comprising 35,804 instances.
\newline\textbf{Catastrophic failure rates.}
The right panel of Figure~\ref{fig:a-mono-spectrum} quantifies the severity of mismatch.
{For each primitive and each $a_{\mathrm{mono}}$ bin, we compute the fraction of games on which that primitive's AUC exceeds the 75th-percentile threshold of all game-primitive AUC. Failure patterns are non-uniform across primitives and across game feature values such as monotonicity.}
The results show substantial regime-dependent degradation in primitive performance across the game spectrum.

\subsection{Game space has solver-aligned latent structure}
\label{subsec:exp-latent}

The mismatch results above motivate the search for a latent coordinate system that explains \emph{why} different solvers succeed in different regimes. 
\newline\textbf{Diagnostic-level winner partitioning.}
Figure~\ref{fig:winner-heatmaps} provides a complementary view on the full 35{,}804-game corpus in the $z_{\mathrm{harm}} \times z_{\mathrm{pot}}$ and  $z_{\mathrm{harm}} \times z_{\mathrm{mono}}$ plane.
A structured partitioning emerges, with GDA and Extragradient primitives each dominating in different regions of game feature space.
\newline\textbf{From fragmented to geometric.}
Figure~\ref{fig:intro-cartography} compares PCA projections of the raw diagnostic space and the learned representation space. In the learned representation, games with similar solver behaviour cluster more coherently.

\subsection{Structure-aware synthesis closes the oracle gap}
\label{subsec:exp-synthesis}

We evaluate the learned solver against baselines that isolate each component, then dissect the structure of the gap to oracle in the learned latent $\hat z$.
\begin{table}[t]
\centering
\caption{Summary results on the 3D payoff benchmark. Gap Closure ($(\mathrm{AUC}_{\text{best-fixed}} - \mathrm{AUC}_{\text{method}}) / (\mathrm{AUC}_{\text{best-fixed}} - \mathrm{AUC}_{\text{oracle}})$) measures the fraction of the best-fixed-to-oracle gap that each method recovers.
}
\label{tab:summary_3d_payoff}
\begin{tabular}{lccc}
\toprule
\textbf{Method} & \textbf{Val AUC} \(\downarrow\) & \textbf{Val Final}  \(\downarrow\) & \textbf{Gap Closure} \\
\midrule
Per-game oracle         & 0.0273 & 0.00251 & 100\% \\
\textbf{Learned (soft)}& \textbf{0.0291}$\pm$1e-4 & 0.0043 $\pm$1e-5 & \textbf{79.3\%} \\
Learned (top-1)          & 0.0295 $\pm$6e-4 & 0.0051 $\pm$2e-6 & 74.7\% \\
Best fixed (GDA)        & 0.0360 & 0.0053 & 0\% \\
Equal weight            & 0.0486 & 0.0152 & $<$0\% \\
\bottomrule
\end{tabular}
\vspace{-4mm}
\end{table}

\paragraph{Performance ladder.}
Table~\ref{tab:summary_3d_payoff} shows a clear performance hierarchy. Equal-weight mixtures underperform the best fixed primitive, confirming that naïve averaging dilutes effective solver behaviour. Structure-conditioned routing substantially closes the oracle gap, while soft mixtures consistently outperform hard top-1 selection, particularly in low-margin transition regions where no single primitive clearly dominates. Remaining oracle advantage concentrates primarily at the structural periphery, where representation fidelity is lower. On a 1,000-game validation subset, the learned soft mixture outperforms every constituent primitive on 3.2\% of games, indicating that convex combinations can capture transition regimes more effectively than hard solver selection.

\paragraph{Latent geometry organises games by solver regime.}
To verify that the recogniser learns a \emph{solver-relevant} representation, we visualise the 5-D $\hat z$ cloud under 2-D PCA, coloured by the per-game best primitive (Figure~\ref{fig:zhat_solver_cluster}). The seven primitive regions are spatially separated and largely contiguous in $\hat z$, despite the fact that the same primitives are not separable in the raw 5-D diagnostic space (left panel). This indicates that the recogniser has discovered structure beyond classical decomposition coordinates, geometrically aligned with which primitive the per-game oracle ultimately prefers. Decomposing the same cloud by oracle-relative outcome (Appendix Figure) further shows that \emph{tie} games, where the learned mixture matches the oracle within tolerance, occupy a compact lobe of the latent space, while games where the oracle retains an advantage scatter along the periphery of each primitive region. The recogniser has therefore picked up the \emph{location} of the residual oracle gap, even though no oracle label was supplied at training time.

\paragraph{Where does the residual oracle gap come from?}
We probe the residual gap with a 5-fold linear classifier on $\hat z$, trained to discriminate ``ours-better'' from ``oracle-better'' games. Globally, the probe attains $\mathrm{AUC}=0.81$, which shows a sizeable fraction of the games where the oracle out-performs the learned solver are in fact linearly separable in $\hat z$. Crucially, when the same probe is fit \emph{within each oracle-best primitive region} (i.e.\ conditioned on the primitive identity), the within-region AUC reaches $0.70$--$0.92$. Two consequences follow. (i)~The information needed to identify oracle-advantaged games is present in $\hat z$. (ii)~Pushing the same probe through the existing softmax policy head, we recover $\mathrm{AUC}\!\approx\!0.81$ on $\log\pi$, so that we can find the discriminative signal is not destroyed by the routing head.

\subsection{Ablation Studies}
\label{sec:ablations}


\begin{table}[t]
\centering
\caption{
Module-wise ablation of the representation pipeline. Lower validation AUC is better.
The ablations separately test the training distribution, input information sources, and representation pathway.
}
\label{tab:ablation}
\setlength{\tabcolsep}{3.5pt}
\renewcommand{\arraystretch}{1.12}
\small
\begin{tabularx}{\linewidth}{lcccc c X}
\toprule
\textbf{Condition}
& \textbf{Pay.}
& \textbf{Diag.}
& \textbf{Rec.}
& \textbf{Sampler}
& \textbf{Val AUC} \(\downarrow\)
& \textbf{Main test} \\
\midrule

\multicolumn{7}{l}{\textit{Reference and training distribution}} \\
Full features
& \(\checkmark\)
& \(\checkmark\)
& \(\checkmark\)
& U
& \(\sim 0.033\)
& Reference pipeline \\

+ prioritised sampling
& \(\checkmark\)
& \(\checkmark\)
& \(\checkmark\)
& P
& \(0.029 \pm 10^{-4}\)
& Hard-game reweighting \\

\midrule
\multicolumn{7}{l}{\textit{Input feature ablations}} \\
Zero diagnostics
& \(\checkmark\)
& \(\times\)
& \(\checkmark\)
& U
& \(0.035 \pm 10^{-4}\)
& Are diagnostics necessary? \\

Zero payoff
& \(\times\)
& \(\checkmark\)
& \(\checkmark\)
& U
& \(0.055 \pm 4{\times}10^{-4}\)
& Is payoff structure necessary? \\

\midrule
\multicolumn{7}{l}{\textit{Representation pathway ablation}} \\
Direct diagnostics
& \(\times\)
& \(\checkmark\)
& \(\times\)
& U
& \(0.047 \pm 2.3{\times}10^{-3}\)
& Can raw diagnostics replace the recogniser? \\

\bottomrule
\end{tabularx}

\vspace{0.3em}
\begin{flushleft}
\footnotesize
\textit{Notes.} Pay. = payoff features; Diag. = optimisation diagnostics; Rec. = recogniser; U = uniform sampling; P = prioritised sampling.
\end{flushleft}
\end{table}

To quantify the contribution of different input features to routing quality, table~\ref{tab:ablation} studies which information sources and representation pathways are necessary for the learned solver. We ablate the 23-dimensional recogniser input and the sampling scheme. 

First, adding priority sampling improved the AUC by 11.5\%. This mechanism fundamentally reshapes the reinforcement learning exploration dynamics. Our preliminary analysis revealed a severe long-tail challenge within the baseline: a mere 5\% of the games accounted for 43\% of the total AUC gap. By dynamically up-weighting gamescharacterised by higher empirical difficulty or larger training losses, priority sampling actively directs the RL agent's exploratory bandwidth toward the most complex and structurally informative regions of the game space. This targeted exploration effectively overcomes the sparse learning signals in hard-to-solve instances, breaking the routing performance bottleneck.
Regarding the input representations, ablating the 5D diagnostics while retaining the 18D payoff features induces only a marginal performance degradation, confirming that the raw payoff matrix encapsulates the primary signal required for solver routing. Conversely, discarding the payoff features to rely exclusively on the 5D diagnostics causes a severe performance drop, demonstrating that classical decomposition coordinates alone are insufficient for accurate routing. Furthermore, when we compare the \emph{zero diagnostics} ablation (where raw payoffs are mapped through the recogniser to a learned representation $\hat{z}$) and the \emph{direct diagnostics} baseline (which feeds raw 5D diagnostics directly to the policy). While both architectures supply a 5D input to the routing policy, the learned representation outperforms the classical diagnostic coordinates by a relative $34\%$. This result confirms that the structure recogniser extracts a task-specific transformation uniquely optimised for routing decisions.

\subsection{Residual correction as a diagnostic instrument}
\label{subsec:exp-residual}

\textbf{Spatial continuity of residual improvements.} We train the residual corrector over the frozen base model via truncated Backpropagation Through Time (BPTT) (window size 5), applying a dense per-step exploitability loss gated by $q_\Phi(G) \in [0,1]$ and regularised by a relative trust penalty. As anticipated, the base model's difficulty surface (AUC prior to residual correction) is spatially continuous within the learned representation space $\hat{z}$. Crucially, the \emph{improvement rate} (the local fraction of games where the residual successfully reduces AUC) exhibits strong spatial coherence (Moran's $I$ = 0.71). This confirms that the efficacy of the residual is fundamentally governed by the inferred game structure rather than stochastic variations.
\newline\textbf{Concentration at solver region boundaries.} Games benefiting from residual correction (Figure~\ref{fig:residual}c) densely cluster at the decision boundaries between solver regions. These are regions characterised by high routing uncertainty, where no pure primitive strictly dominates. 
This conditional activation demonstrates that the residual effectively compensates for local inadequacies in the primitive hull while remaining appropriately dormant where the base mechanisms suffice. Importantly, the observed spatial coherence ensures the residual targets geometrically distinct regions where convex interpolation falls short. Boundary-localised activation suggests missing intermediate primitives, whereas peripheral activation may indicate insufficient representation resolution.
\newline\textbf{Residuals as structural diagnostics.} The topological distribution of high-improvement clusters in $\hat{z}$ space serves as an interpretable diagnostic tool for evaluating the primitive library. Clusters situated at the interfaces between solver regions (e.g., the transition zone between extragradient and GDA regimes) indicate an algorithmic gap, suggesting the need for an intermediate mechanism tailored to these crossover dynamics. Meanwhile, clusters at the extreme periphery of the latent manifold imply resolution limits within the structure representation itself, where structurally disparate games are erroneously mapped to adjacent coordinates. Consequently, persistent high-activation clusters provide targeted candidates for discovering novel primitive mechanisms or refining the resolution of the representation space.

\section{Discussion and Limitations}
\label{sec:discussion}

The learned representation is solver-aligned rather than a canonical game decomposition, and its coordinates need not admit direct game-theoretic interpretation. Empirically, hard solver identities are often unstable outside high-margin regimes, supporting the use of soft mixtures in transition regions. The residual acts only as a bounded local corrector, and the induced cartography should be interpreted as an empirical geometry of solver behaviour rather than a complete classification of games. More broadly, the results suggest that solver-relevant structure is richer than existing analytical coordinates.

\section{Conclusion}
\label{sec:conclusion}

We introduced a structure-aware framework for synthesising game solvers from primitive update mechanisms and bounded residual corrections. The method learns a map from realised game structure to solver behaviour, using a recogniser to construct solver-relevant coordinates and a policy to form soft mixtures over primitive dynamics. A gated residual provides controlled local corrections beyond the primitive hull.
The central message is not that game space admits a simple hard partition into solver classes. Instead, our results support a margin-aware algorithmic cartography: high-margin regions reveal interpretable solver regions, while low-margin transition regions favour soft mixtures over hard primitive selection. This perspective reframes equilibrium computation as the joint problem of learning adaptive solver mixtures and mapping the geometry of solvability. It opens a path toward data-driven discovery of where classical algorithms succeed, where they fail, and how their mechanisms can be combined.

\bibliographystyle{abbrvnat}
\bibliography{sample}

\clearpage
\appendix

\section{Further Results on Game--Solver Geometry}
\label{app:further_results}

This appendix provides additional evidence for the central claim of the paper: solver behaviour induces a meaningful geometry over games, and the learned recogniser maps this geometry into a latent space that supports structure-conditioned solver synthesis. In this section, first, we show that oracle-best primitives form spatially coherent territories in diagnostic game space. Second, we show that the recogniser maps these game regions into solver-aligned regions in the learned \(\hat z\) space. Third, we examine transition games where hard primitive selection is insufficient and soft mixtures outperform every constituent primitive. Finally, we use the residual diagnostic map to show where the primitive mixture still has systematic blind spots.

\subsection{Robustness on a Broader Payoff Benchmark}
\label{app:broad10d_results}

Table~\ref{tab:broad10dim_t60_summary} repeats the summary comparison on the 10D payoff benchmark. The same qualitative hierarchy as in Table~\ref{tab:summary_3d_payoff} is preserved. We can find that the learned soft mixture substantially closes the gap between the best fixed primitive and the per-game oracle, while hard top-1 selection is considerably weaker and equal weighting performs worse than the best fixed primitive. This supports our interpretation that the learned policy is not merely memorising a low-dimensional benchmark, but is exploiting solver-relevant structure that remains useful in a broader payoff space.

\begin{table}[t]
\centering
\caption{Summary results on the 10D payoff benchmark. Gap Closure uses the same definition as Table~\ref{tab:summary_3d_payoff}.}
\label{tab:broad10dim_t60_summary}
\begin{tabular}{lccc}
\toprule
\textbf{Method} & \textbf{Val AUC} \(\downarrow\) & \textbf{Val Final} \(\downarrow\) & \textbf{Gap Closure} \\
\midrule
Per-game oracle            & 0.02512 & 0.00696 & 100.0\% \\
\textbf{Learned (soft)}    & \textbf{0.02771}$\pm$3e-4 & 0.00863$\pm$1e-5 & \textbf{66.1\%} \\
Learned (top-1)            & 0.02986$\pm$8e-4 & 0.01044$\pm$1e-5 & 37.9\% \\
Best fixed (extragradient) & 0.03275 & 0.01040 & 0.0\% \\
Equal weight               & 0.05614 & 0.03237 & $<$0\% \\
\bottomrule
\end{tabular}
\vspace{-4mm}
\end{table}

\subsection{Oracle Solver Territories in Game Space}
\label{app:oracle_solver_territories}

\begin{figure}[t]
    \centering
    \includegraphics[width=\linewidth]{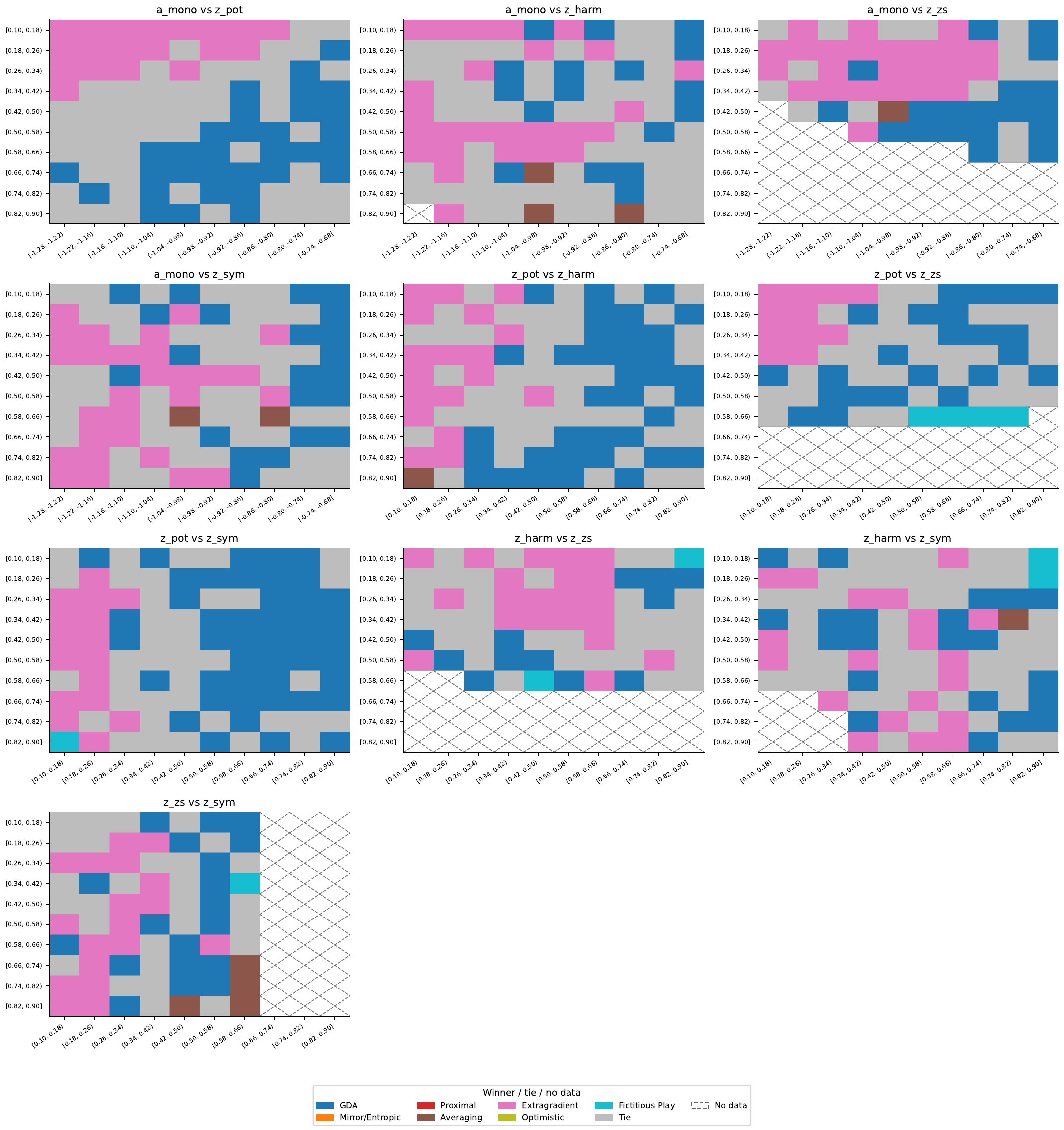}
    \caption{
    Oracle-best primitive across pairwise diagnostic game features. Each panel bins games by two diagnostic coordinates and colours each cell by the primitive with the lowest validation AUC in that bin. Grey cells denote statistical ties, where no primitive is significantly better than the others. The main pattern is spatial coherence: primitive dominance forms contiguous territories rather than random patches.
    }
    \label{fig:win_primitives}
\end{figure}

We first ask whether different primitives dominate in different parts of the game distribution, or whether the identity of the best primitive is essentially random. Figure~\ref{fig:win_primitives} summarises the oracle-best primitive across pairwise diagnostic planes. Each sub-panel bins games by two diagnostic coordinates and colours each bin by the primitive with the best validation AUC in that region. Grey cells indicate statistical ties, where the performance gap between primitives is below the significance threshold.

The important observation is that the winning primitive varies smoothly across diagnostic space. Large contiguous regions are dominated by GDA and extragradient, while averaging and fictitious play appear in more localised regimes. This spatial organisation suggests that solver optimality is governed by continuous structural coordinates rather than by isolated game identities. It also motivates a soft solver-selection mechanism: near territory boundaries, several primitives may be nearly competitive, and a hard winner-take-all selection can be too brittle.

\subsection{Pure-Solver Sensitivity in the Learned Latent Space}
\label{app:pure_solver_sensitivity}

Winner maps identify which primitive is best in each region, but they discard the magnitude of solver mismatch. To examine whether the learned latent space carries a continuous performance geometry, we visualise the oracle-relative retention score of each pure primitive,
\[
r_i(G)
=
\frac{\mathrm{AUC}_{\mathrm{best}}(G)+\varepsilon}
{\mathrm{AUC}(g_i;G)+\varepsilon}.
\]
This score is the reciprocal of the oracle-normalised rollout loss used for training, specialised to a pure primitive. Values close to one indicate that primitive \(g_i\) is near the per-game best pure solver, while smaller values indicate stronger degradation.

\begin{figure}[t]
    \centering
    \includegraphics[width=\linewidth]{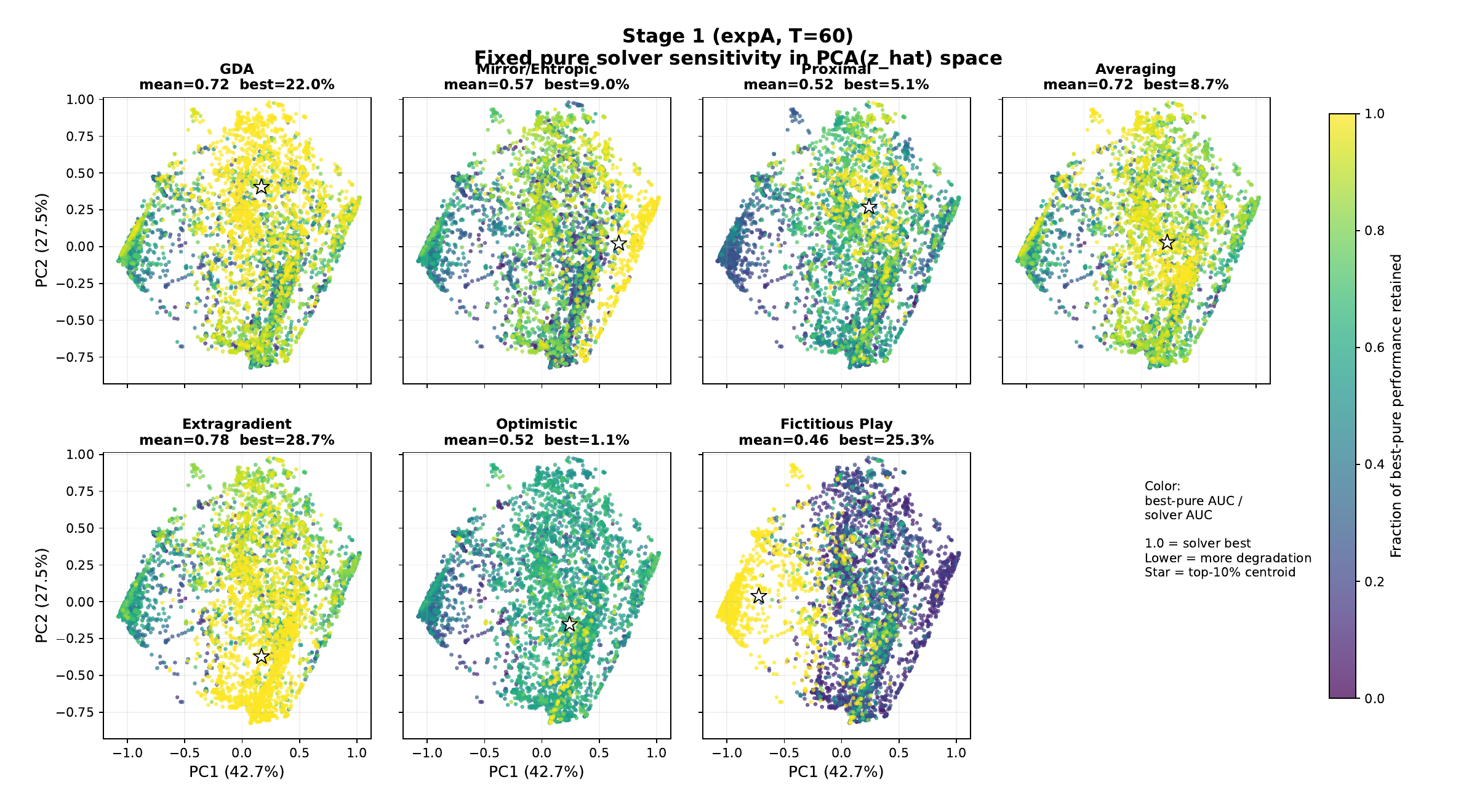}
    \caption{
        Pure-solver sensitivity in learned \(\hat z\) space. Each panel shows one pure primitive over the same PCA projection of \(\hat z\). Point colour denotes the oracle-relative retention score. The spatially coherent performance patterns show that \(\hat z\) encodes continuous solver-performance sensitivity.
        }
    \label{fig:pure_solver_sensitivity_zhat}
\end{figure}

Figure~\ref{fig:pure_solver_sensitivity_zhat} shows that pure-solver sensitivity varies smoothly over PCA\((\hat z)\) space. Different primitives achieve high retention in different regions of the latent manifold, and their degradation patterns are spatially coherent. This gives direct geometric meaning to \(\hat z\): nearby games in the learned representation tend to induce similar relative performance profiles across pure solvers. The transition bands between high-retention regions correspond to the regimes where hard primitive selection is brittle and soft mixtures are most useful.

\subsection{Learned Primitive-Weight Landscapes}
\label{app:primitive_weight_landscapes}

\begin{figure}[t]
    \centering
    \includegraphics[width=\linewidth]{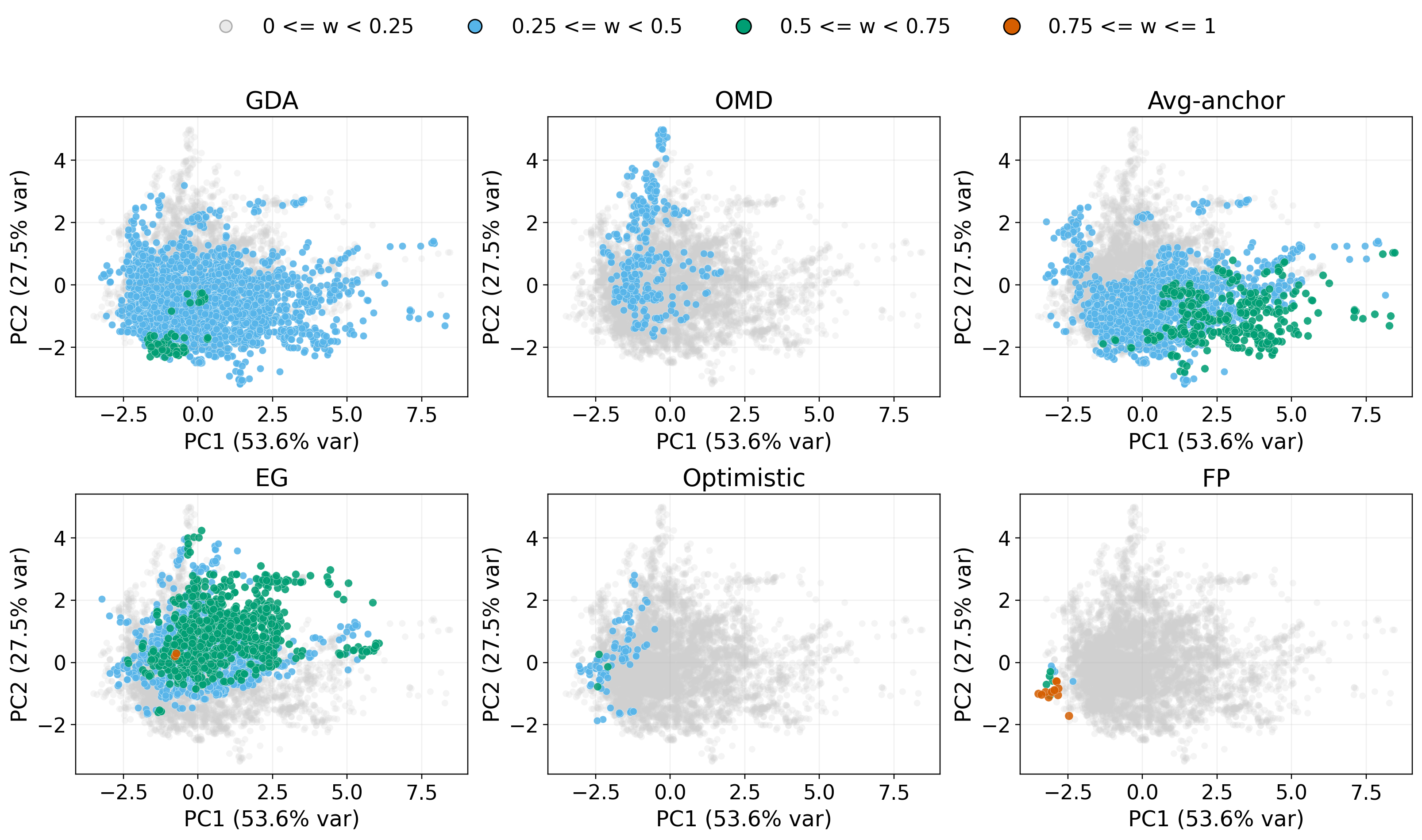}
    \caption{
    Learned primitive-weight landscapes in PCA\((\hat z)\) space. Each panel shows the same validation games projected onto the first two principal components of the learned recogniser representation \(\hat z\). The panel title indicates the primitive, and point colour denotes the mixture coefficient assigned to that primitive by the policy head. Grey points correspond to near-zero or small weights, while blue, green, and orange indicate progressively larger coefficients.
    }
    \label{fig:primitive_weight_landscapes}
\end{figure}

The previous analysis shows that pure-solver performance varies smoothly over the learned \(\hat z\) manifold. We next ask whether the learned policy head uses this geometry in a similarly structured way. Figure~\ref{fig:primitive_weight_landscapes} visualises the mixture coefficient assigned to each primitive over the same PCA projection of \(\hat z\). Each panel corresponds to one primitive, and point colour indicates the magnitude of the learned mixture weight.

The weight maps are spatially organised. GDA receives broad moderate weight over a large central region; OMD and optimistic updates are activated only in more localised parts of the manifold; averaging and extragradient occupy different but partially overlapping regions; and fictitious play is concentrated in a small peripheral regime. This confirms that the policy head is not simply using a fixed global mixture. Instead, it learns a soft routing landscape over \(\hat z\), assigning different primitives to different regions while retaining overlap in transition zones.

This figure complements the pure-solver sensitivity plots. Whereas Figure~\ref{fig:pure_solver_sensitivity_zhat} shows where each primitive is individually effective, Figure~\ref{fig:primitive_weight_landscapes} shows how the learned solver converts that latent performance geometry into mixture coefficients. The overlap between non-zero weight regions provides direct visual evidence that the learned solver performs soft synthesis rather than hard primitive selection.

\subsection{From Game Features to Solver-Aligned Latent Geometry}
\label{app:zhat_solver_mapping}

\begin{figure}[t]
    \centering
    \includegraphics[width=\linewidth]{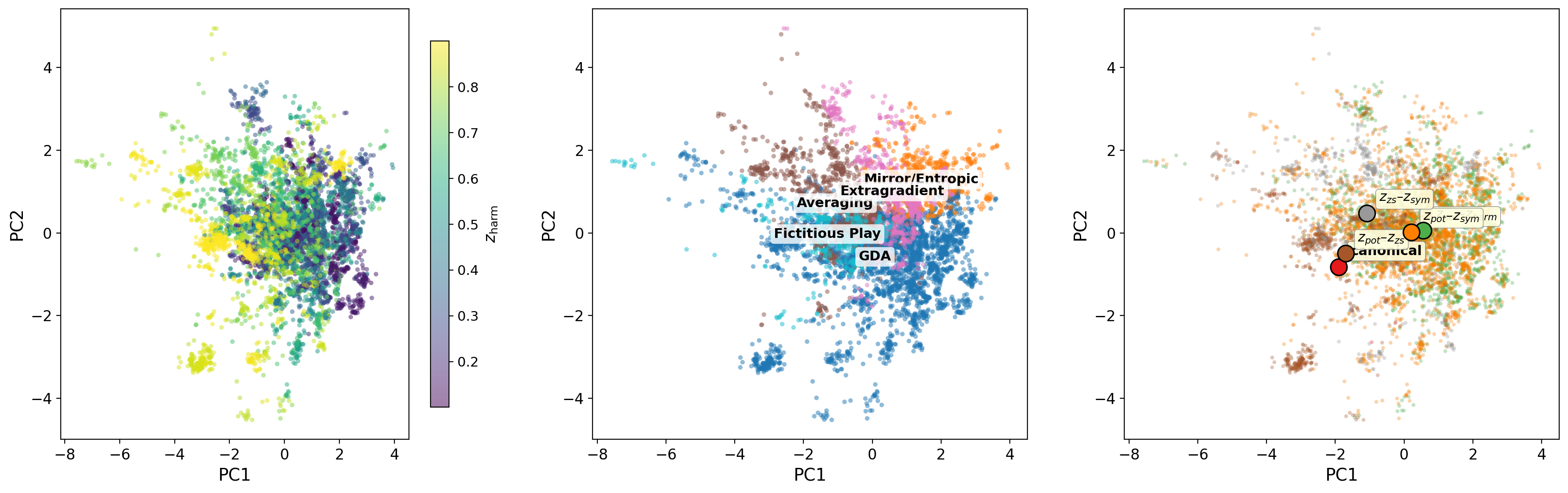}
    \caption{
    Mapping from game features to learned \(\hat z\) and solver clusters. The figure from left to right: raw game features provide the original structural coordinates; the recogniser maps these games into the learned \(\hat z\) space; the policy head then assigns solver regions or dominant primitive clusters.
    }
    \label{fig:zhat_solver_cluster}
\end{figure}

Having established that oracle solver choice has spatial structure in game-feature space, we next examine whether the learned recogniser captures this structure. Figure~\ref{fig:zhat_solver_cluster} visualises the mapping from game features to the learned latent representation \(\hat z\), and then to solver cluster assignments. The purpose of this figure is to check whether \(\hat z\) is merely a compressed representation, or whether it reorganises games into solver-aligned territories.

The mapping shows that games with similar solver behaviour are placed near each other in the learned latent space. In other words, the recogniser does not simply preserve raw diagnostic similarity; it reshapes the input into a geometry that is more directly aligned with primitive selection. This supports the interpretation of \(\hat z\) as a solver-facing game representation: it is learned to expose the structural information needed by the policy head to route between primitives.

\subsection{Inside Each Primitive Territory: Where Does the Oracle Still Win?}
\label{app:per_primitive_winloss_regions}

The previous subsections established two coarse-grained facts: primitives form contiguous territories in game space (Figure~\ref{fig:win_primitives}), and the recogniser maps those territories into solver-aligned regions of \(\hat z\) (Figure~\ref{fig:zhat_solver_cluster}). A further question is: \emph{within} each primitive's \(\hat z\) territory, where do the games that the per-game oracle still beats the learned mixture actually live, and is that location systematic?

\begin{figure}[t]
    \centering
    \includegraphics[width=\linewidth]{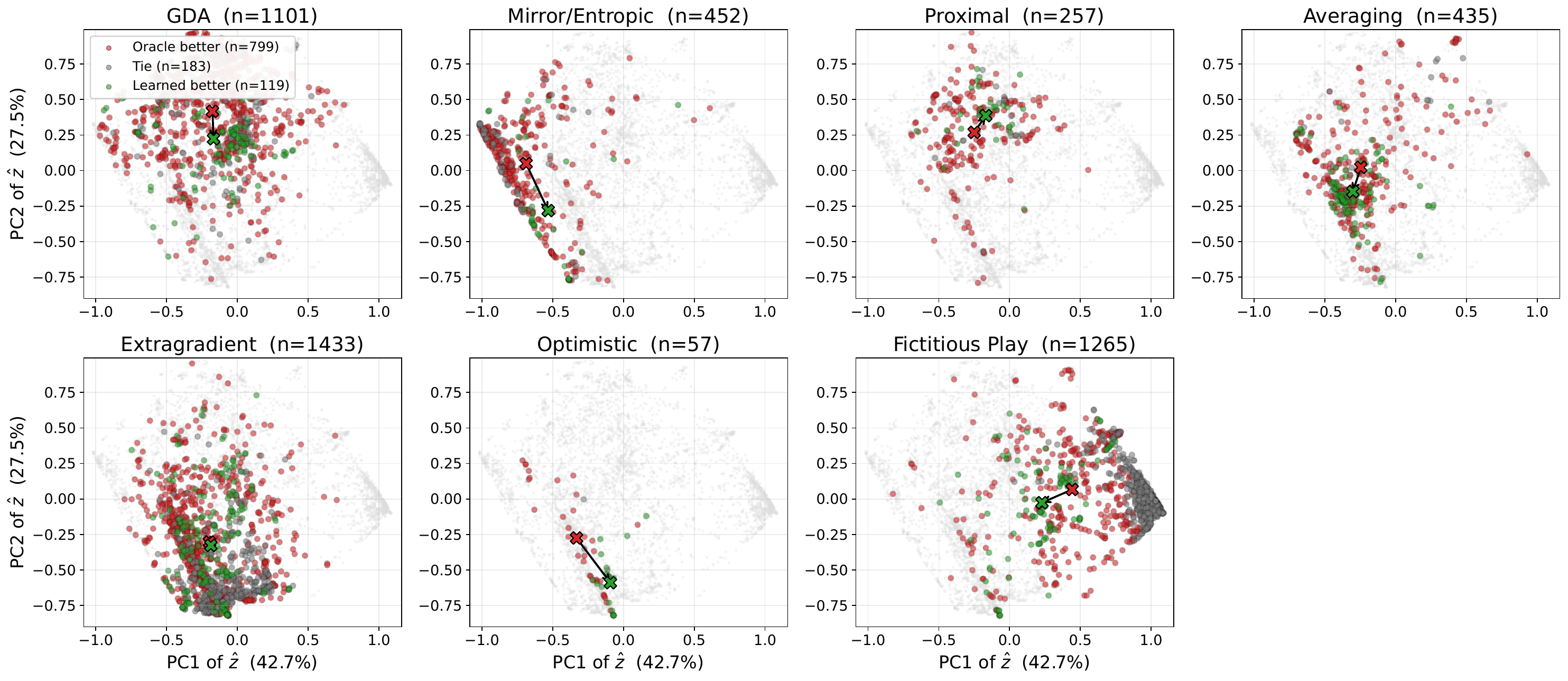}
    \caption{
    Per-oracle-primitive win/tie/loss regions in shared PCA(\(\hat z\)) coordinates (\(N=5{,}000\) validation games). Each panel keeps games whose per-game oracle-best primitive is the named one; remaining games are drawn faintly in grey for spatial reference. Within each territory: red dots are games where the per-game oracle still beats the learned mixture, green dots are games where the learned mixture is at least as good, and grey dots are ties. The red and green \(\times\) markers are bucket centroids, and the black arrow points from the oracle-better centroid to the learned-better centroid.
    }
    \label{fig:per_primitive_winloss_regions}
\end{figure}

Figure~\ref{fig:per_primitive_winloss_regions} addresses this directly. For each oracle-best primitive, we colour the games in shared PCA(\(\hat z\)) coordinates by their oracle-relative outcome: oracle-better (red), tie (grey), and learned-better (green); games outside the focal primitive are drawn faintly for spatial reference. One structural observations stand out.

\textbf{Win and loss regions are spatially separated within each territory.} In every primitive panel, the red and green clouds occupy different sub-regions of the same territory rather than mixing uniformly. The arrows connect the oracle-better centroid (red \(\times\)) to the learned-better centroid (green \(\times\)) and have non-trivial length in all seven panels. This means the recogniser already encodes, locally inside each primitive's region, \emph{where in the manifold} the residual oracle gap concentrates (this is consistent with the within-region linear-probe AUC of \(0.70\)--\(0.92\) reported in Section~\ref{subsec:exp-synthesis}).


\subsection{Soft Mixtures Help on Transition Games}
\label{app:mixture_transition_games}

The territorial structure in Figure~\ref{fig:win_primitives} also explains why convex mixtures can outperform hard primitive selection. If a game lies near a boundary between solver territories, no single primitive may cleanly capture the best update mechanism. In such transition regimes, the optimal update direction can require interpolation between primitive behaviours.

\begin{figure}[h]
    \centering
    \includegraphics[width=0.8\linewidth]{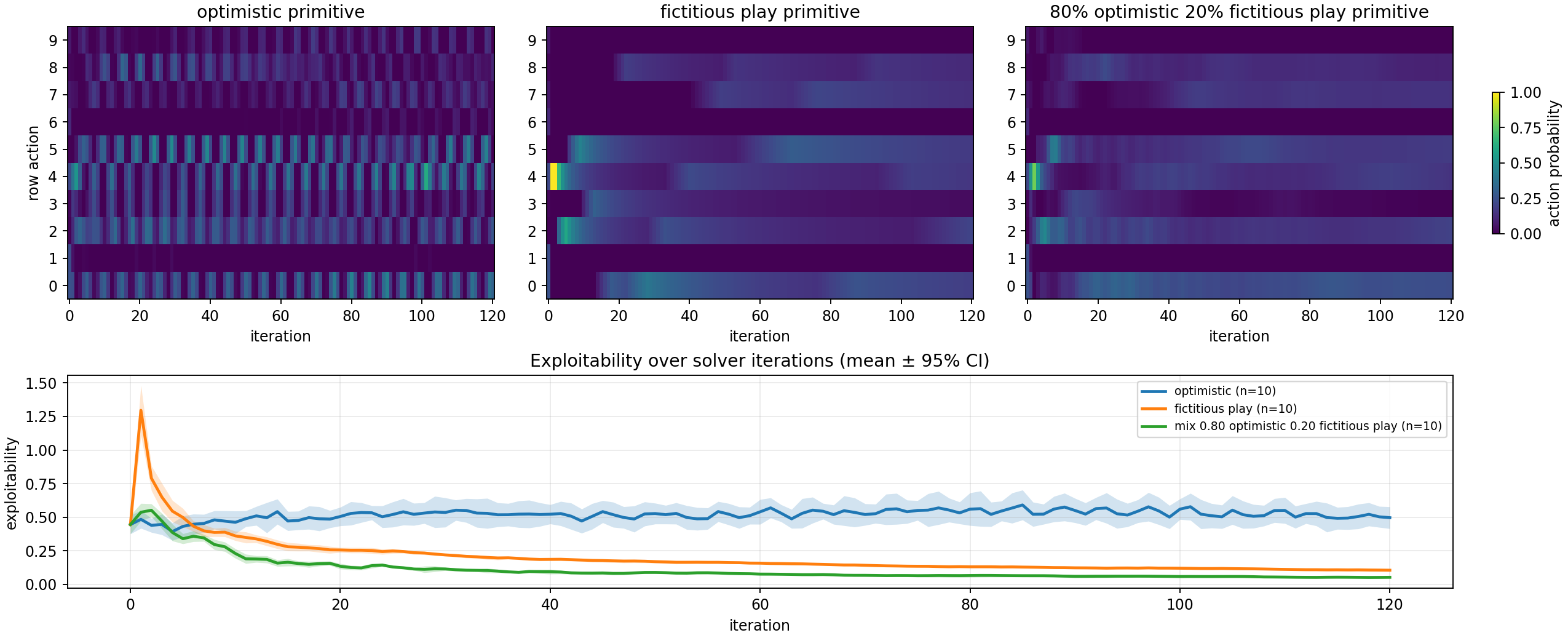}
    \caption{
    Example transition game where the learned convex mixture outperforms its constituent primitives. The game lies in a region where no single primitive cleanly dominates, so hard primitive selection is suboptimal. The learned mixture interpolates between solver mechanisms and achieves lower exploitability than each individual primitive.
    }
    \label{fig:mixture}
    \vspace{-6pt}
\end{figure}

Figure~\ref{fig:mixture} shows a representative transition game where the learned convex mixture achieves lower exploitability than any of its constituent primitives individually. This is important because it demonstrates that the learned policy is not only selecting among primitives; it is synthesising a new update direction by combining them. The result provides per-instance evidence for the value of soft solver synthesis, especially in structural regimes where the diagnostic map does not assign a clear unique winner.

\subsection{Residuals Reveal Boundary Blind Spots}
\label{app:residual_diagnostic}

\begin{figure}[t]
    \centering
    \includegraphics[width=\linewidth]{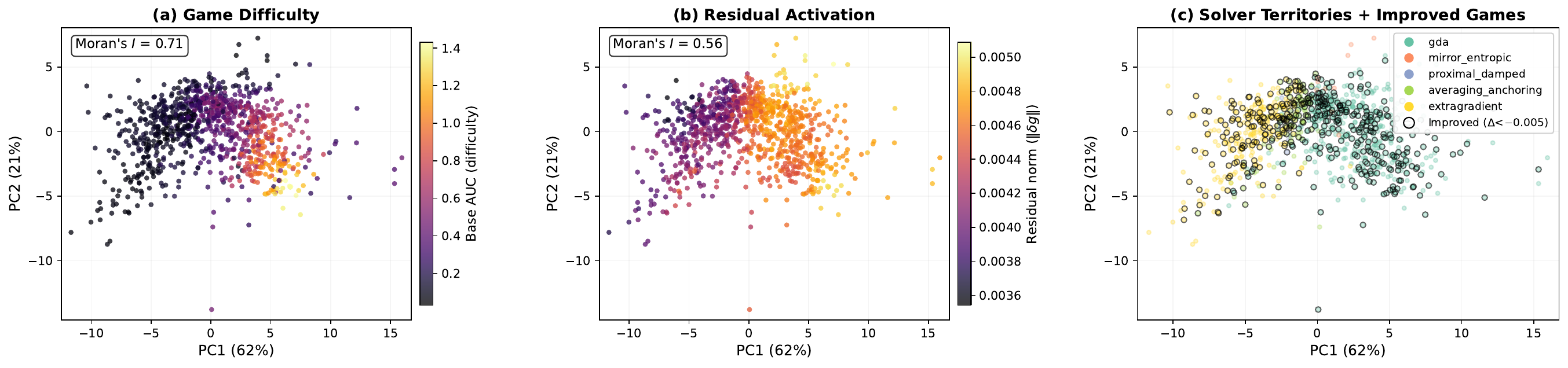}
    \caption{
    Residual diagnostic map in \(\hat z\) space over \(N=2{,}000\) probe games. 
    \textbf{(a)} Base game difficulty without residual correction, measured by validation AUC. Spatial continuity, measured by Moran's \(I=0.71\), shows that difficult games form coherent regions in latent space. 
    \textbf{(b)} Residual activation intensity. The residual response is also spatially coherent, with Moran's \(I=0.56\), indicating that the residual reacts to structured regions rather than random games. 
    \textbf{(c)} Solver regions coloured by dominant primitive, with residual-improved games overlaid as orange circles \((\Delta\mathrm{AUC} < -0.005, n=423)\). Improvements concentrate near territory boundaries, suggesting that the residual mainly corrects local blind spots of the primitive mixture where routing uncertainty is highest.
    }
    \label{fig:residual}
    \vspace{-4mm}
\end{figure}

Finally, we examine where the residual correction is useful. If the primitive mixture already spans all useful update directions, residual activation should be diffuse or unnecessary. Instead, Figure~\ref{fig:residual} shows that both game difficulty and residual activation are spatially coherent in \(\hat z\) space. This indicates that the residual is not acting as arbitrary noise; it responds to structured regions of the learned game manifold.

Panel (a) shows base difficulty without residual correction. The high Moran's \(I\) indicates that difficult games cluster in latent space rather than appearing randomly. Panel (b) shows residual activation intensity, which is also spatially coherent. Panel (c) overlays games improved by the residual on top of solver regions. The improved games concentrate near territory boundaries, where routing certainty is lowest and where the primitive basis is most likely to be incomplete. This supports the interpretation of the residual as a local corrector for boundary cases, rather than as a replacement for the primitive mixture.

\paragraph{Summary.}
Together, these supplementary results support the solver-cartography view of the method. Oracle-best primitives form coherent territories in game-feature space; the recogniser maps these territories into a solver-aligned latent geometry; soft mixtures improve performance on transition games where hard selection is insufficient; and the residual correction is most useful near territory boundaries. The appendix therefore complements the main experiments by showing not only that the method performs well, but also where and why each component is useful.

\clearpage
\section{Algorithm}

Algorithm~\ref{alg:solver-synthesis} summarises the solver.

\begin{algorithm}[H]
\caption{Structure-aware solver synthesis with static primitive mixture}
\label{alg:solver-synthesis}
\KwIn{game \(G=(A,B)\); primitive basis \(\{g_i\}_{i=1}^M\); recogniser and policy parameters \(\Phi\); step sizes}
Initialise \(x_0 \in \Delta^n\) and \(y_0 \in \Delta^m\)\;
Compute the structure representation \(\hat z=h_\Phi(A,B)\)\;
Compute primitive logits \(r_\Phi(\hat z)\)\;
Compute the soft primitive mixture \(w_{\mathrm{soft}}=\mathrm{softmax}(\beta r_\Phi(\hat z))\)\;
Optionally compute the hard top-1 primitive \(w_{\mathrm{hard}}=e_{\arg\max_i r_{\Phi,i}(\hat z)}\)\;
\For{\(t=1,\dots,T\)}{
Compute primitive solver updates \(\{g_i(\theta_t)\}_{i=1}^M\)\;
Construct \(g_t^{\mathrm{prior}}=\sum_i w_i g_i(\theta_t)\)\;
Compute diagnostics \(u_t\) for analysis and residual geometry\;
Build residual geometry features \(\mathcal S_t\)\;
Compute gated residual correction \(q_\Phi(G)\delta g_t\)\;
Construct \(g_t^{\mathrm{synth}}=g_t^{\mathrm{prior}}+q_\Phi(G)\delta g_t\)\;
Apply the projected solver update to obtain \(\theta_{t+1}\)\;
}
\end{algorithm}

\section{Joint training objective}
\label{sec:joint-training}

\subsection{Complete solver synthesis pipeline}
\label{sec:solver-synthesis}

We now describe the full solver synthesis mechanism. At each iteration, the method combines a global, decomposition-informed representation of the game with local optimisation diagnostics in order to construct an adaptive update rule.


The solver architecture consists of four stages:
(i) a structure recogniser producing a solver-aligned representation,
(ii) a structure-conditioned primitive mixture,
(iii) a bounded residual correction module,
and (iv) a projected update step.
Equations~(7)--(14) provide the complete computational pipeline used throughout the paper.

\paragraph{Optimisation diagnostics.}
To capture the local optimisation regime, we define
\begin{equation}
u_t
=
\big(
\mathrm{gap}_t,\;
\|g_t\|,\;
\mathrm{rot}_t,\;
\mathrm{align}_t,\;
\mathrm{step}_t,\;
\mathrm{sym\_loc}_t,\;
\mathrm{skew\_loc}_t,\;
t/T
\big),
\label{eq:optimisation-diagnostics}
\end{equation}
where the components capture equilibrium violation,
gradient magnitude, rotational structure,
update alignment, local Jacobian symmetry,
and optimisation phase.


\paragraph{Primitive mixture.}
Primitive mixture.
Let $\{g_i\}_{i=1}^M$ denote the primitive solver library.
The structure-conditioned prior update is

\begin{equation}
g_t^{\mathrm{prior}}
=
\sum_{i=1}^M
w_i(\hat z;\Phi)\,
g_i(\theta_t),
\qquad
w_i \ge 0,
\qquad
\sum_i w_i = 1.
\label{eq:primitive-mixture-app}
\end{equation}

\paragraph{Residual correction.}

To permit local deviations beyond the primitive hull,
a bounded residual module produces
\begin{equation}
\delta g_t
=
\lambda_{\max}
\tanh(f_\Phi(S_t)),
\label{eq:residual-app}
\end{equation}
where $S_t$ contains optimisation diagnostics,
primitive statistics, and local trajectory information. The final synthesised update direction is
\begin{equation}
g_t^{\mathrm{synth}}
=
g_t^{\mathrm{prior}}
+
q_\Phi(G)\delta g_t.
\label{eq:synth-update-app}
\end{equation}

\paragraph{Trust-region penalty.}
The residual is regularised through the relative trust penalty
\begin{equation}
R_{\mathrm{trust}}
=
\lambda_{\mathrm{trust}}
\sum_{t=1}^T
\frac{
\|\delta g_t\|^2
}{
\|g_t^{\mathrm{prior}}\|^2 + \varepsilon
}.
\label{eq:trust-penalty-app}
\end{equation}


\paragraph{Projected update rule.}
Let
\(
g_t^{\mathrm{synth}}
=
(g_{x,t}^{\mathrm{synth}},
g_{y,t}^{\mathrm{synth}})
\).
The projected update is
\begin{align}
x_{t+1}
&=
\Pi_{\Delta_n}
\Big(
x_t
+
\eta_t g_{x,t}^{\mathrm{synth}}
\Big),
\\
y_{t+1}
&=
\Pi_{\Delta_m}
\Big(
y_t
+
\eta_t g_{y,t}^{\mathrm{synth}}
\Big).
\label{eq:projected-update-app}
\end{align}



\subsection{Training dynamics}
\label{subsec:exp-training}
\paragraph{Training procedure.}
Optimisation proceeds in two stages.
Phase~I (routing initialisation) trains both the structure recogniser and routing policy via KL divergence against oracle one-hot targets, with no differentiable rollout.
The policy temperature anneals from $0.5$ to ${\approx}0.13$ over 15 epochs.
During Phase I, routing accuracy improves substantially,
with behavioural KL decreasing from $1.29$ to $0.49$.
Phase~II (end-to-end rollout optimisation) loads the best Phase~I checkpoint and applies differentiable rollout-based AUC minimisation.

Furthermore, empirical analysis reveals that the performance gap relative to the best pure oracle is highly concentrated under uniform sampling, with just 5\% of games accounting for 43\% of the total AUC gap. This concentration motivates  the implementation of loss-based prioritised sampling to systematically redirect gradient updates toward the most structurally challenging instances. Ultimately, this end-to-end refinement yields a 12.54\% relative improvement in validation performance, reducing the mean AUC exploitability from $0.0311$ to $0.0272$.



\section{Structural Diagnostics}
\label{sec:diagnostics_definition}
The following diagnostics provide interpretable structural probes used
for analysis, dataset construction, and comparison against the learned
solver-aligned representation.
They are not assumed to constitute a complete coordinate system for solvability.

Game features, or game structural diagnostics, are defined below:

\paragraph{Potential coordinate \(z_{\mathrm{pot}}\).}
For \(i,i' \in [n]\) and \(j,j' \in [n]\), define the row and column cross-differences
\[
\Delta^A_{ii'jj'} = A_c(i,j)-A_c(i',j)-A_c(i,j')+A_c(i',j'),
\]
\[
\Delta^B_{ii'jj'} = B_c(i,j)-B_c(i,j')-B_c(i',j)+B_c(i',j').
\]
The normalised potential residual is
\[
\mathrm{gap}_{\mathrm{pot}}
=
\frac{\Bigl(\sum_{i,i',j,j'} (\Delta^A_{ii'jj'}-\Delta^B_{ii'jj'})^2\Bigr)^{1/2}}
{\Bigl(\sum_{i,i',j,j'} \bigl((\Delta^A_{ii'jj'})^2+(\Delta^B_{ii'jj'})^2\bigr)\Bigr)^{1/2}+\varepsilon}.
\]
We then define
\[
z_{\mathrm{pot}} = \max\{0,\, 1-\mathrm{gap}_{\mathrm{pot}}\}.
\]
Hence \(z_{\mathrm{pot}}\approx 1\) indicates a game close to exact-potential structure.

\paragraph{Harmonic coordinate \(z_{\mathrm{harm}}\).}
Define the zero-sum component
\[
Z = \frac{1}{2}(A_c-B_c),
\]
and its skew-symmetric part
\[
H = \frac{1}{2}(Z-Z^\top).
\]
The harmonic strength is \(\|H\|_F\). Let
\[
D = \|A_c\|_F + \|B_c\|_F + \varepsilon.
\]
Then
\[
z_{\mathrm{harm}} = \min\Bigl\{1,\; \frac{2\|H\|_F}{D}\Bigr\}.
\]
Thus \(z_{\mathrm{harm}}\) measures cyclic/rotational content.

\paragraph{Zero-sum coordinate \(z_{\mathrm{zs}}\).}
Define thenormalised zero-sum residual
\[
\mathrm{gap}_{\mathrm{zs}} = \frac{\|A_c+B_c\|_F}{\|A_c\|_F+\|B_c\|_F+\varepsilon},
\]
and set
\[
z_{\mathrm{zs}} = \max\{0,\, 1-\mathrm{gap}_{\mathrm{zs}}\}.
\]
Hence \(z_{\mathrm{zs}}\approx 1\) means the game is close to zero-sum.

\paragraph{Symmetry coordinate \(z_{\mathrm{sym}}\).}
Define thenormalised symmetry residual
\[
\mathrm{gap}_{\mathrm{sym}} = \frac{\|A_c-B_c^\top\|_F}{\|A_c\|_F+\|B_c\|_F+\varepsilon},
\]
and set
\[
z_{\mathrm{sym}} = \max\{0,\, 1-\mathrm{gap}_{\mathrm{sym}}\}.
\]
Thus \(z_{\mathrm{sym}}\approx 1\) means the game is close to player-symmetric.

\paragraph{Monotonicity score \(a_{\mathrm{mono}}\).}
Let the game Jacobian be
\[
J =
\begin{bmatrix}
0 & -A \\
-B^\top & 0
\end{bmatrix}.
\]
Its symmetric part is
\[
S = \frac{J+J^\top}{2}
=
\frac{1}{2}
\begin{bmatrix}
0 & -(A+B) \\
-(A^\top+B^\top) & 0
\end{bmatrix}.
\]
We define
\[
a_{\mathrm{mono}} = \lambda_{\min}(S).
\]
Positive values indicate strong monotonicity, values near zero indicate the monotone boundary, and negative values indicate non-monotonicity.
%

\section{Residual as a probe of missing structure}\label{app:residual_probe}

Residual activation provides a diagnostic signal for representation
or primitive mismatch. If \(\|\delta g(G)\|\) is persistently large over a subset \(\mathcal S \subset \mathcal G\), then the primitive basis and the current structure coordinates are insufficient to explain solver behaviour there. Such regions suggest either missing primitives, missing structure coordinates, or interactions not captured by the current representation.
%
%
%
A relative trust-region penalty~\cite{schulman2015trust}:
\begin{align}
\mathcal{R}_{\mathrm{trust}}
=
\lambda_{\mathrm{trust}}
\sum_{t=1}^T
\frac{\|\delta g_t\|^2}{\|g^{\mathrm{prior}}_t\|^2 + \varepsilon}
\end{align}
ensures that the residual remains a controlled perturbation rather than replacing the primitive mixture. The tanh bounding and trust penalty together guarantee that $\|\delta g_t\| \le \lambda_{\max}$ at all times.

\section{Dataset Construction}
\label{sec:data_construction}
\paragraph{Dataset construction.}
Our dataset is built to cover the diagnostic space defined in \ref{sec:diagnostics_definition}
\[
(z_{\mathrm{pot}}, z_{\mathrm{harm}}, z_{\mathrm{zs}}, z_{\mathrm{sym}}, a_{\mathrm{mono}}).
\]
At a high level, we sample games from a collection of canonical generators, compute their diagnostic coordinates, and retain candidates by rejection sampling so that occupied bins in low-dimensional diagnostic projections are as uniformly populated as possible. We then perform targeted upsampling in under-covered regions of the feature space until the resulting balanced pairwise slices are as complete as possible, followed by deduplication of repeated games. Thus the final dataset is not an i.i.d.\ sample from a single game distribution; rather, it is a coverage-oriented corpus designed to populate the structural feature space.

Accordingly, empirical results should be interpreted as evaluating
solver behaviour across a broad structural coverage distribution
rather than estimating performance under a natural game prior.

The corpus comprises 35{,}804 two-player $3\times 3$ games spanning zero-sum, potential, harmonic, symmetric, and interpolated regimes (generation details in Appendix~\ref{sec:data_construction}). We split the corpus into 28{,}643 training and 7{,}161 validation games (80/20) via a single deterministic shuffle. All reported validation metrics are computed on the held-out 20\% partition.

\paragraph{Centred and normalised payoffs.}
Let \(\tilde A,\tilde B \in \mathbb{R}^{n\times n}\) denote the raw payoff matrices. We first center each player's payoffs,
\[
\bar A = \frac{1}{n^2}\sum_{i,j}\tilde A_{ij},
\qquad
\bar B = \frac{1}{n^2}\sum_{i,j}\tilde B_{ij},
\]
\[
A^\circ = \tilde A - \bar A \mathbf{1}\mathbf{1}^\top,
\qquad
B^\circ = \tilde B - \bar B \mathbf{1}\mathbf{1}^\top.
\]
We then apply a shared normalization
\[
s = \max\!\Bigl\{\max_{i,j}|A^\circ_{ij}|,\ \max_{i,j}|B^\circ_{ij}|,\ 10^{-8}\Bigr\},
\]
and store
\[
A = \frac{A^\circ}{s},
\qquad
B = \frac{B^\circ}{s}.
\]
Thus each generated game has zero-mean payoffs for each player and a common max-entry scale.

\section{Primitive Solvers}
\label{sec:primitive_solvers_definition}
\begin{table}[h]
\centering
\small
\begin{tabular}{ll}
\toprule
Primitive & Intended regime \\
\midrule
GDA & potential-like / monotone \\
Mirror/Entropic & simplex-constrained optimisation \\
Extragradient & rotational / adversarial \\
Optimistic & cycling correction \\
Averaging & stabilised interpolation regimes \\
Fictitious play & response-dominant regimes \\
\bottomrule
\end{tabular}
\caption{Primitive solver library and intended optimisation regimes.}
\end{table}
Let \(x \in \Delta^{n-1}\) and \(y \in \Delta^{m-1}\) denote the row- and column-player mixed strategies, and let
\[
g_x(y) = Ay,
\qquad
g_y(x) = B^\top x.
\]
Each primitive updates \((x,y)\) and projects back to the simplex when needed. Below, \(\Pi_\Delta\) denotes Euclidean projection onto the simplex, \(\eta>0\) is a step size, and \(t\) is the iteration index.

\paragraph{Gradient Descent--Ascent (GDA).}
\[
x^{t+1} = \Pi_\Delta\!\bigl(x^t + \eta g_x(y^t)\bigr),
\qquad
y^{t+1} = \Pi_\Delta\!\bigl(y^t + \eta g_y(x^t)\bigr).
\]

\paragraph{Mirror/Entropic.}
The update is multiplicative:
\[
x^{t+1}_i \propto x^t_i \exp\!\bigl(\eta\, g_x(y^t)_i\bigr),
\qquad
y^{t+1}_j \propto y^t_j \exp\!\bigl(\eta\, g_y(x^t)_j\bigr),
\]
followed by normalization to the simplex.

\paragraph{Proximal (damped).}
First form the GDA proposal
\[
\tilde x^{t+1} = \Pi_\Delta(x^t + \eta g_x(y^t)),
\qquad
\tilde y^{t+1} = \Pi_\Delta(y^t + \eta g_y(x^t)),
\]
then damp it with parameter \(\rho \in (0,1]\):
\[
x^{t+1} = \Pi_\Delta\bigl((1-\rho)x^t + \rho \tilde x^{t+1}\bigr),
\qquad
y^{t+1} = \Pi_\Delta\bigl((1-\rho)y^t + \rho \tilde y^{t+1}\bigr).
\]

\paragraph{Averaging / anchoring.}
Let \(\bar x^t,\bar y^t\) be anchor strategies, initialised at the uniform strategy and updated by running averages. With anchor strength \(\gamma \ge 0\),
\[
x^{t+1} = \Pi_\Delta\bigl(x^t + \eta g_x(y^t) + \gamma(\bar x^t - x^t)\bigr),
\]
\[
y^{t+1} = \Pi_\Delta\bigl(y^t + \eta g_y(x^t) + \gamma(\bar y^t - y^t)\bigr),
\]
and then
\[
\bar x^{t+1} = \frac{t}{t+1}\bar x^t + \frac{1}{t+1}x^{t+1},
\qquad
\bar y^{t+1} = \frac{t}{t+1}\bar y^t + \frac{1}{t+1}y^{t+1}.
\]

\paragraph{Extragradient.}
Compute the extrapolated point
\[
x^{t+\frac12} = \Pi_\Delta(x^t + \eta g_x(y^t)),
\qquad
y^{t+\frac12} = \Pi_\Delta(y^t + \eta g_y(x^t)),
\]
then evaluate the gradient there:
\[
x^{t+1} = \Pi_\Delta\bigl(x^t + \eta g_x(y^{t+\frac12})\bigr),
\qquad
y^{t+1} = \Pi_\Delta\bigl(y^t + \eta g_y(x^{t+\frac12})\bigr).
\]

\paragraph{Optimistic gradient.}
Let \(g_x^{t-1}, g_y^{t-1}\) denote the previous gradients. Then
\[
x^{t+1} = \Pi_\Delta\bigl(x^t + \eta(2g_x(y^t)-g_x^{t-1})\bigr),
\]
\[
y^{t+1} = \Pi_\Delta\bigl(y^t + \eta(2g_y(x^t)-g_y^{t-1})\bigr).
\]

\paragraph{Fictitious play.}
Let \(\bar x^t,\bar y^t\) denote the empirical averages. Define best responses
\[
\mathrm{BR}_x(\bar y^t) \in \arg\max_{x \in \Delta^{n-1}} x^\top A \bar y^t,
\qquad
\mathrm{BR}_y(\bar x^t) \in \arg\max_{y \in \Delta^{m-1}} (\bar x^t)^\top B y.
\]
Then fictitious play updates the empirical averages by
\[
x^{t+1} = \frac{t}{t+1}x^t + \frac{1}{t+1}\mathrm{BR}_x(y^t),
\qquad
y^{t+1} = \frac{t}{t+1}y^t + \frac{1}{t+1}\mathrm{BR}_y(x^t).
\]
In the implementation, the best responses are taken against the running averages and the state stores those averages explicitly.

\clearpage
\section{Primitive Behaviours}

\begin{table*}[b]
\centering
\caption{Mapping between primitive solver mechanisms and convergence properties across game classes in normal-form games. Results are representative and summarise well-established findings in the learning-in-games literature.}
\label{tab:primitive-convergence-map}
\footnotesize
\setlength{\tabcolsep}{4pt}

\begin{tabularx}{\textwidth}{@{} >{\bfseries\raggedright\arraybackslash}p{2.8cm} >{\raggedright\arraybackslash}X >{\raggedright\arraybackslash}X >{\raggedright\arraybackslash}X @{}}
\toprule
\textbf{Primitive / Solver Family} 
& \textbf{Game Class} 
& \textbf{Convergence Guarantee} 
& \textbf{Limitations / Failure Modes} \\
\midrule

Gradient / descent dynamics
& Potential games; identical-interest games; concave–convex optimisation problems
& Converges to Nash equilibrium (often globally) under smoothness and step-size conditions
& Fails in non-potential games; exhibits cycling in zero-sum or rotational (skew-symmetric) components
\\ \addlinespace[0.6em]

Mirror descent / entropic dynamics
& Convex–concave zero-sum games; regularised games; monotone variational inequality settings
& Converges to Nash equilibrium under monotonicity; improves stability via geometry of the simplex
& May exhibit limit cycles in non-monotone or strongly rotational games; convergence rates degrade outside structured regimes
\\ \addlinespace[0.6em]

Extra-gradient / Mirror-Prox
& Monotone games; bilinear zero-sum games; saddle-point problems
& Converges to Nash equilibrium with last-iterate or ergodic guarantees under monotonicity
& Requires monotonicity or near-monotonicity; may be conservative or inefficient in potential-dominated regimes
\\ \addlinespace[0.6em]

Optimistic gradient methods
& Zero-sum and near-zero-sum games; weakly monotone operators
& Stabilises cycling and achieves convergence in certain bilinear and smooth saddle-point problems
& Can diverge or oscillate in non-rotational or non-adversarial regimes; sensitive to step-size tuning
\\ \addlinespace[0.6em]

Best-response dynamics
& Potential games; dominance-solvable games; acyclic games
& Converges to pure-strategy Nash equilibrium under improvement-path conditions
& May cycle in general games; convergence depends on update order and tie-breaking rules
\\ \addlinespace[0.6em]

Better-response / logit dynamics
& Potential games; stochastic perturbations of best-response dynamics
& Converges to (approximate) Nash equilibrium or stationary distributions concentrated on equilibria
& May not converge to exact Nash equilibrium; behaviour depends on noise level and temperature parameter
\\ \addlinespace[0.6em]

Fictitious play
& Zero-sum games; \(2\times 2\) games; potential games
& Time-averaged strategies converge to Nash equilibrium in zero-sum games; pointwise convergence in some structured classes
& Fails to converge in general non-zero-sum games; exhibits cycling (e.g.\ Shapley examples)
\\ \addlinespace[0.6em]

Smoothed / stochastic fictitious play
& Broader class of games including some non-potential games
& Converges to Nash equilibrium under smoothing and regularity assumptions
& Requires carefully tuned perturbations; convergence may be slow or sensitive to noise
\\ \addlinespace[0.6em]

No-regret dynamics (e.g.\ regret matching)
& General finite normal-form games
& Empirical play converges to correlated equilibrium (CE) or coarse correlated equilibrium (CCE)
& Does not guarantee convergence to Nash equilibrium; may converge to distributions far from Nash
\\ 
\bottomrule
\end{tabularx}
\end{table*}

\end{document}